\documentclass[10pt,journal,compsoc]{IEEEtran}
\ifCLASSOPTIONcompsoc
\usepackage[nocompress]{cite}
\else
\usepackage{cite}
\fi

\usepackage{amsmath}
\usepackage{amssymb}
\usepackage{amsthm}
\usepackage{multirow}
\usepackage{array}
\usepackage{bm}
\usepackage{enumerate}
\usepackage{mathrsfs}

\usepackage{booktabs}
\usepackage{graphicx}
\usepackage{adjustbox}
\usepackage{relsize}

\usepackage{algorithmic}
\usepackage{algorithm}

\usepackage{url}
\usepackage{bbm}
\usepackage{dsfont}
\usepackage{array}
\usepackage{courier}
\usepackage{bbding}

\usepackage{color}
\usepackage{subfigure}

\usepackage{makecell}
\usepackage{enumitem}
\usepackage{orcidlink}

\theoremstyle{remark}

\begin{document}
	
	\title{Deep Isolation Forest for Anomaly Detection}

	%
	%
	
	\author{
		Hongzuo~Xu\orcidlink{0000-0001-8074-1244},
		Guansong~Pang\orcidlink{0000-0002-9877-2716},
		Yijie~Wang\orcidlink{0000-0002-2913-4016}
		and~Yongjun~Wang\orcidlink{0000-0002-3257-539X}
		\IEEEcompsocitemizethanks{
			\IEEEcompsocthanksitem Hongzuo Xu, Yijie Wang and Yongjun Wang are with the College of Computer, National University of Defense Technology, Changsha 410073, PR China. Hongzuo Xu and Yijie Wang are also with the Science and Technology on Parallel and Distributed Processing Laboratory. \protect\\
			E-mail: \{xuhongzuo13, wangyijie, wangyongjun\}@ nudt.edu.cn
			\IEEEcompsocthanksitem Guansong Pang is with the School of Computing and Information Systems, Singapore Management University Singapore, 178902, Singapore. \protect\\
			E-mail: gspang@smu.edu.sg
			\IEEEcompsocthanksitem Guansong Pang and Yijie Wang are corresponding authors. 
		}
	}
	
	%
	%

\markboth{
	 IEEE TRANSACTIONS ON KNOWLEDGE AND DATA ENGINEERING
}%
{
	 Xu \MakeLowercase{\textit{et al.}}: Deep Isolation Forest for Anomaly Detection
}

%



\IEEEtitleabstractindextext{%
	\begin{abstract}
		
		Isolation forest (iForest) has been emerging as arguably the most popular anomaly detector in recent years due to its general effectiveness across different benchmarks and strong scalability. Nevertheless, its linear axis-parallel isolation method often leads to (i) failure in detecting hard anomalies that are difficult to isolate in high-dimensional/non-linear-separable data space, and (ii) notorious algorithmic bias that assigns unexpectedly lower anomaly scores to artefact regions. These issues contribute to high false negative errors. Several iForest extensions are introduced, but they essentially still employ shallow, linear data partition, restricting their power in isolating true anomalies. Therefore, this paper proposes deep isolation forest. We introduce a new representation scheme that utilises casually initialised neural networks to map original data into random representation ensembles, where random axis-parallel cuts are subsequently applied to perform the data partition. This representation scheme facilitates high freedom of the partition in the original data space (equivalent to non-linear partition on subspaces of varying sizes), encouraging a unique synergy between random representations and random partition-based isolation. Extensive experiments show that our model achieves significant improvement over state-of-the-art isolation-based methods and deep detectors on tabular, graph and time series datasets; our model also inherits desired scalability from iForest. 
	\end{abstract}
	
	}

\maketitle

\IEEEdisplaynontitleabstractindextext

%
\IEEEpeerreviewmaketitle

\IEEEraisesectionheading{\section{Introduction}\label{sec:introduction}}

%
%
%

\IEEEPARstart{A}{nomaly} detection has broad applications in various domains, such as detection of insurance fraud and financial crime, surveillance of complex systems like data centres and spacecraft, and identification of attacks and potential threats in cyberspace. 
Given these important applications, this task has been a popular research topic for decades, and numerous anomaly detection approaches have been introduced \cite{aggarwal2017outlieranalysis,pang2021survey}.

In recent years, isolation forest (iForest) \cite{liu2008isolation} has been emerging as arguably the most popular anomaly detector due to its general effectiveness across different benchmarks and strong scalability. 
Compared to many existing methods such as distance/density-based methods, iForest better captures the key essence of anomalies, i.e., ``few and different''. It does not introduce extra assumptions of data characteristics, thus showing consistently effective performance across diverse datasets.
Also, iForest is with linear time complexity, which is often a very appealing advantage in many industrial applications when there are large-scale data and strict requirements of time efficiency. 
Concretely, iForest uses an ensemble of isolation trees (iTree), in which each iTree is grown by iteratively branching. Leaf nodes are built by using random cuts in the values of randomly selected features until the data objects are isolated. Data abnormality is estimated according to the average depth traversing from the root node to the isolated leaf node in these iTrees.

\begin{figure}[t]
	\centering
	\includegraphics[width=0.48\textwidth]{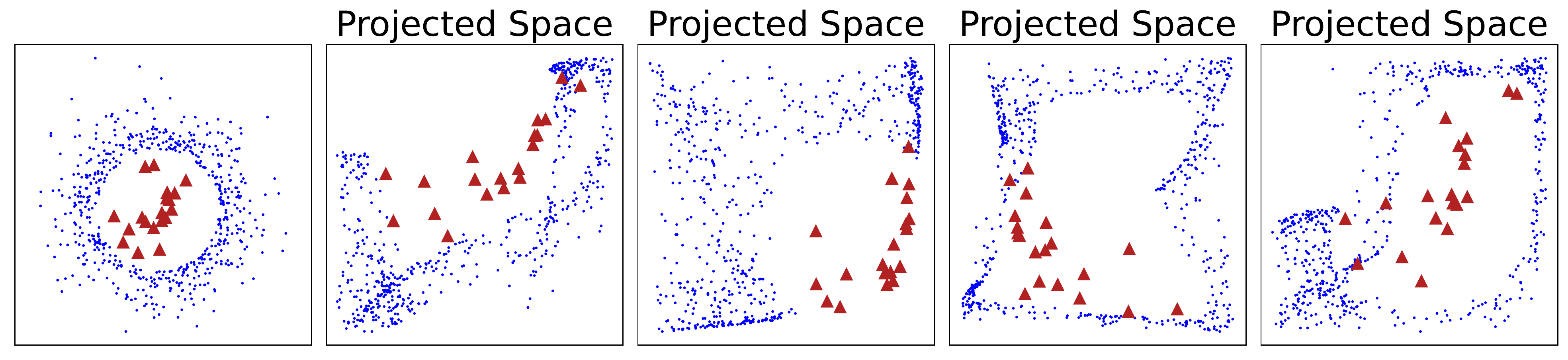}
	\caption{Illustration of the hard anomaly challenge. The left figure is a synthetic dataset, where red triangles are anomalies and blue points are normal samples. The following figures are projected spaces (i.e., random representations) created by our method. 
		Linear data partition (in either vertical/horizontal or oblique forms) used in existing methods cannot effectively isolate these hard anomalies in the original data space. By contrast, these anomalies are possible to be exposed in newly created spaces. 
	}
	\label{fig:hardanom}
\end{figure}

Nevertheless, an explicit major issue is that it cannot handle hard anomalies (e.g., the anomalies that can be only isolated in higher-order subspaces by looking into the combination of multiple features) because it treats all features separately and considers only one feature per isolation operation. 
Fig. \ref{fig:hardanom} exemplifies this issue with a simple toy example. Anomalies (represented as red triangles) are surrounded by ring-shaped normal samples, which cannot be isolated by either x-axis or y-axis slicing cuts.
Although these anomalies might be finally isolated by multiple cuts, it results in indistinguishable isolation depth in iTrees compared to normal data. 
The failure of recalling these anomalies induces high false negative errors. 
Given that anomalies often contain critical information related to potential accidents or faults, those false negatives may cause serious consequences. 
Therefore, this issue has been a major bottleneck hindering the performance of iForest, particularly on datasets with high-dimensional/non-linear-separable data spaces.

\begin{figure}[t]
	\centering
	\includegraphics[width=0.48\textwidth]{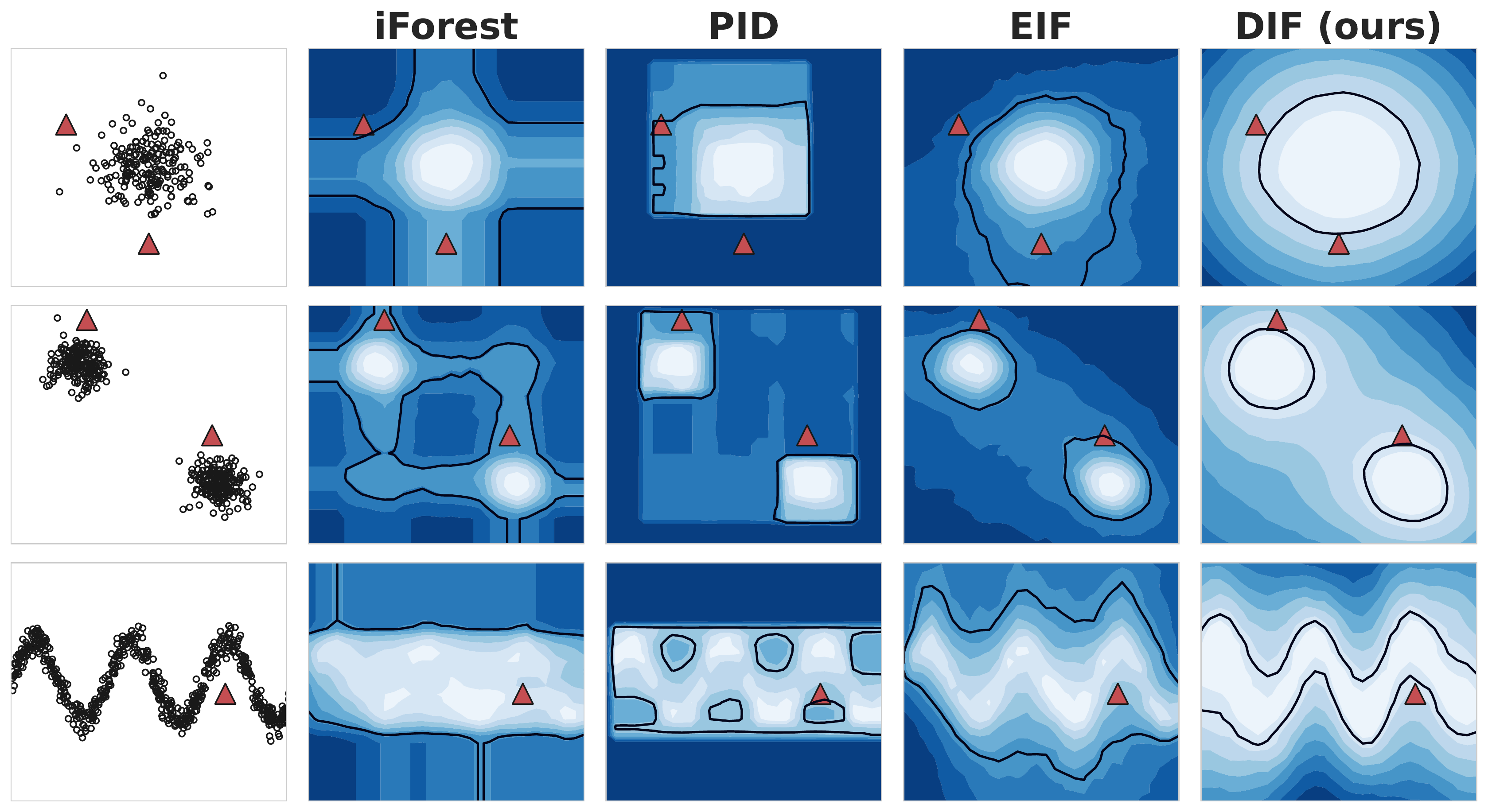}
	\caption{Illustration of the algorithmic bias problem.
		The three figures on the far left are three synthetic datasets, and the following panels of each dataset are the anomaly score maps produced by iForest and its two state-of-the-art extensions -- PID and EIF -- and our method DIF. Score maps indicate anomaly score distribution in the full data space (deeper colour indicates higher abnormality). 
		Black contour lines in each score map denote the 99th percentile of the anomaly scores of the original data, which can be seen as the boundary of data normality predicted by each anomaly detector. 
		Red triangles represent possible anomalies.
		There are some artefact regions in the score maps of iForest and its two extensions, and these methods may fail to identify the anomalies since these anomalies are assigned similarly low anomaly scores as the presented normal samples (they are included in the predicted normal areas). By contrast, DIF produces more accurate and smooth score maps, and DIF can successfully assign a high abnormal degree to anomalies compared to those original data samples.
	}
	\label{fig:teaser}
\end{figure}

Another inherent imperfection of iForest is that it assigns unexpectedly low anomaly scores to artefacts introduced by the algorithm itself, which is revealed as the ``ghost region'' problem in \cite{hariri2019eif}. 
To clearly demonstrate this issue, we visualise the data distribution of three 2-D synthetic datasets used in \cite{hariri2019eif} and anomaly score maps generated by iForest in the first two columns in Fig. \ref{fig:teaser}, respectively. As can be seen from all three anomaly score maps, iForest assigns clearly lower anomaly scores to some artefact regions, i.e., the four rectangular areas centred around the presented data objects in the single-blob dataset (1st row), the upper right and bottom left clique areas in the two-blob dataset (2nd row), and the vertical rectangular areas along the sinusoid in the sinusoidal dataset (3rd row). However, these artefact regions are similar to other regions that contain no samples or have similar radial distances to the presented data objects.
The unusually lower anomaly scores are due to the intrinsic algorithmic bias of iForest, i.e., only axis-parallel partitions are admitted in iTree construction. 
Instead, an expected anomaly score map should smoothly approximate circular contour lines w.r.t. the density of the presented data. 
This problem also leads to false negative errors, i.e., 
iForest may fail to detect possible anomalies in these ``ghost regions'' such as the red triangles in Fig. \ref{fig:teaser} as they are assigned similarly low anomaly scores as some of the presented normal objects and included in the normal areas.

There have been a number of extensions of iForest introduced over the years 
\cite{liu2010detecting,hariri2019eif,gopalan2019pid,lesouple2021generalized,cortes2021revisiting,tokovarov2022probabilistic}. 
These extensions attempt to devise more advanced isolation methods by (i) using novel hierarchical criteria to selectively pick splitting thresholds and/or dimensions \cite{gopalan2019pid,tokovarov2022probabilistic,cortes2021revisiting} or (ii) using optimal/random hyper-planes \cite{liu2010detecting,hariri2019eif,lesouple2021generalized}.
These improvements successfully contribute to better detection performance, but their isolation methods are still confined by linear partition.
Instead, data partition in the isolation process is expected to form non-linearly arbitrary shapes in the data space to handle hard anomalies in complicated datasets and eliminate the algorithmic bias in derived anomaly scores. 
We show two state-of-the-art extensions (PID \cite{gopalan2019pid} and EIF\cite{hariri2019eif}) in Fig. \ref{fig:teaser}. 
PID \cite{gopalan2019pid} selectively chooses splitting dimensions and uses similar axis-parallel isolation operations. Thus, its score maps also encounter the same algorithmic bias problem. 
EIF \cite{hariri2019eif} works better due to the use of hyper-planes with random slopes and intercepts as branching criteria, eliminating the limits of axis-parallel partitions to some extent. However, the partitions still operate linearly, which works ineffectively on challenging datasets where only non-linear partitions are useful in isolating anomalies from the other data. This problem is reflected by the rather wrinkled anomaly score contours in all three datasets in Fig. \ref{fig:teaser}.

Therefore, the key to these limitations becomes very clear now, i.e., a step further is required to liberate the isolation method from the underlying linear constraints.
To this end, this paper proposes a novel extension of iForest named \textsc{Deep Isolation Forest} (DIF for short). 
The key idea in DIF is to harness the strong representation power of neural networks to map the original data into a group of new data spaces, and non-linear isolation can be easily achieved by performing simple axis-parallel partitions upon these newly created data spaces (equivalent to non-linear partition on subspaces of varying sizes in the original data space).
Specifically, we propose a novel representation scheme -- random representation ensemble -- produced by optimisation-free deep neural networks. These networks are only casually initialised and do not involve any optimisation or training process. 
DIF then simply utilises random axis-parallel cuts upon these representations. 
The randomness in our representation scheme allows high freedom of the partition in the original data space, encouraging a unique synergy effect between random representations and random partition-based isolation. 
This facilitates effective isolation of hard anomalies and eliminates the algorithmic bias, thus significantly enhancing the detection performance.
Fig. \ref{fig:hardanom} shows four random representations created by DIF, where those hard anomalies are possible to be exposed and easily isolated by using a few axis-parallel cuts. 
Besides, as shown in Fig. \ref{fig:teaser}, DIF accurately assigns anomaly scores to the presented data objects and has a smooth estimation of the anomaly scores in the other regions, alleviating the aforementioned algorithmic bias problem.

Further, in our specific implementation of DIF, we propose the Computation-Efficient Representation Ensemble method (CERE) and the Deviation-Enhanced Anomaly Scoring function (DEAS).
With the help of CERE, DIF can efficiently produce the representation ensemble by taking full advantage of parallel accelerators in the mini-batch calculation, largely eliminating the computational overhead. 
DEAS leverages the hidden quantitative information enclosed in the mapped dense representations along with qualitative comparisons. This additional information enables a more accurate assessment of the isolation difficulty of data objects, leading to a better anomaly scoring function.

Our main contributions are summarised as follows:

\begin{itemize}
	\item 
	We propose the \textsc{Deep Isolation Forest} (DIF) method. A new isolation method is introduced, which enables non-linear partition on subspaces of varying sizes, offering a more effective anomaly isolation solution than the current methods that are capable of linear isolation only.
	The resulting approach can better handle hard anomalies in complex datasets with high-dimensional/non-linear-separable data spaces and eliminate the algorithmic bias.
	We also show that DIF is a high-level generalisation of iForest and its very recent extension EIF. 
	
	\item We propose a novel representation scheme, the random representation ensemble, in which only casually initialised neural networks are required. This representation scheme facilitates high freedom of partitions in the original data space.
	The unique synergy between random representations and random partition-based isolation brings excellent randomness and desired diversity into the overall ensemble-based abnormality estimation process.

	\item We propose the representation ensemble method CERE to ensure that DIF can inherit good scalability from iForest. The anomaly scoring function DEAS is introduced to leverage additional quantitative information enclosed in mapped dense representations, enhancing the quality of anomaly scoring of DIF.

	\item DIF offers a data-type-agnostic anomaly detection solution. We show that DIF is versatile to detect anomalies in different types of data by simply plugging in corresponding randomly initialised neural networks in the feature mapping. 
	
\end{itemize}

Extensive experiments on a large collection of real-world datasets, including not only tabular data but graph and time series data, show that: (i) DIF significantly outperforms iForest and its state-of-the-art extensions (Sec. \ref{sec:effectiveness}); (ii) DIF also achieves remarkable improvement compared to the ensemble of advanced deep anomaly detectors (Sec. \ref{sec:effectiveness}); (iii) DIF has desired scalability on high-dimensional, large-scale datasets (Sec. \ref{sec:time}) 
and presents good robustness to anomaly contamination (Sec. \ref{sec:robustness}); (iv) the significance of the synergy between random representations and random partition-based isolation is justified by comparing to several alternatives (Sec. \ref{sec:significance}); 
and (v) the contribution of CERE and DEAS is separately verified in our ablation study (Sec. \ref{sec:ablation}).

\section{Related Work}

\subsection{Anomaly Detection}
Anomaly detection has been intensively studied in the last decades by using different data characteristics like distance, density, clustering membership, or probability \cite{aggarwal2017outlieranalysis}. 
Recent studies \cite{pang2018Learning,ruff2018dsvdd,bergman2019classification,wang2021rdp} devise deep anomaly detection models based on the representation learning, while some other methods use concepts like reconstruction \cite{liu2021rca} or deviation \cite{pang2019deep} in the deep learning framework as scoring function.
Surveys and comparative studies can be found in \cite{pang2021survey,ruff2021unifying,emmott2013systematic}. Also, some practical anomaly detection tools, e.g., \cite{lai2021tods,alnegheimish2022sintel,zhao2019pyod,ibm}, are developed to facilitate the use of anomaly detection models in real-world applications.

\subsection{Isolation Forest and Its Extensions}

iForest \cite{liu2008isolation} is a very popular anomaly detection method that identifies anomalies according to the isolating difficulty in the data space. 
Although iForest has shown effective performance on several benchmarks, a number of its extensions have been proposed, attempting to fix its notorious vulnerabilities and obtain better detection performance.

The mainstream of these extensions focuses on devising more effective isolation methods. 
SCIF \cite{liu2010detecting} introduces a non-axis-parallel way as branching criteria, i.e., instead of selecting one feature during splitting, an optimal slicing hyper-plane is used. EIF \cite{hariri2019eif} also uses hyper-planes but with random slopes and intercepts. 
Another work \cite{lesouple2021generalized} further fixes the empty branching problem of EIF 
by choosing splitting thresholds from the range of projected values onto the slope direction. 
PID \cite{gopalan2019pid} 
selectively picks dimensions that have greater variance and selects split points according to the sparsity of the partition branches. The literature \cite{tokovarov2022probabilistic} proposes a probability-based method to find better split values than random splitting. 
These extensions normally achieve better performance by introducing non-axis-parallel and/or heuristic partition.
However, the major problem is that they still rely on \textit{linear isolation operations}, which means it is also hard to handle complicated data that require non-linear partitions as the isolation method. Also, they suffer from the aforementioned artefact problem due to the implicit algorithmic bias hidden in the isolation process. 
There are also some extensions employing the nearest neighbour information into isolation process, e.g., LeSiNN \cite{pang2015lesinn} and iNNE \cite{bandaragoda2018nearest}. 
The literature \cite{zhang2017lshiforest} further uses locally-sensitive hashing to extend the isolation mechanism to any distance measures and data types.
These distance-based methods introduce an extra assumption, i.e., anomalies are far from other data objects. However, this assumption does not always hold since clustered anomalies are also very close to their adjacent neighbours \cite{liu2010detecting}. The detection performance is also sensitive to the choice of distance metrics.

Another angle is to enhance the scoring method. The literature \cite{mensi2021enhanced} introduces a path-weighted scoring method and a probability-based aggregation function. PID \cite{gopalan2019pid} redefines the scoring function according to the sparsity rather than depth in the tree. Similarly, these extensions are still vulnerable due to their linear isolation methods.

\subsection{Deep Ensembles}

Deep ensemble \cite{lakshminarayanan2017de}, a simple framework that combines prediction results of a group of independently trained networks together, has garnered much interest.
It can improve prediction accuracy and provide uncertainty estimation in a simple framework without modifying the original working pipeline.
Similarly to other ensemble-based approaches, the quality of deep ensembles also largely hinges on the diversity of its members.
Besides, albeit simple, deep ensembles still induce considerably larger computational costs. 
Thus, many related studies attempt to address these two key limitations, e.g., \cite{angelo2021repulsive} uses repulsive terms to ensure individual diversity, 
\cite{rame2021dice} increases diversity 
by adversarially preventing features from being conditionally predictable from each other,
and \cite{nam2021diversity} proposes a distilled model that can absorb as much function diversity inside the ensemble as possible.
In addition, this framework inspires related studies on anomaly detection. The literature \cite{chen2017outlier,kieu2019outlier} combines a group of autoencoders and uses the median of reconstruction errors as anomaly scores. To improve member diversity, these autoencoders are sparsely connected.

Our work also involves the process of integrating neural networks. From the deep ensemble aspect, our work can effectively tackle the above two key issues. The diversity between ensemble members can be guaranteed and the calculation efficiency can be well maintained since only initialised networks are required.  
Our work may also provides valuable insights into the deep ensemble research line.

\section{Problem Statement and Notations}

Let $\mathcal{D}=\{\bm{o}_1, \cdots, \bm{o}_N \}$ be a dataset with $N$ data objects, anomaly detection is to give a scoring function $f: \mathcal{D} \mapsto \mathbb{R}^{N}$ that estimates the abnormality of each data object. Different from many existing studies that only focus on an individual data type, we do not restrict the type of data objects in this work, which means they can be multi-dimensional vectors, time series data, or graphs.
Throughout the paper, we use calligraphic fonts for sets, script typeface for functions, bold lowercase letters for
vectors, bold uppercase letters to denote matrices. Table \ref{tab:notation} summarises main notations.

\begin{table}[htbp]
	\centering
	\caption{The main notations used in the paper}
	\scalebox{0.9}{
		\begin{tabular}{p{2.6cm}p{6.4cm}}
			\hline
			\textbf{Format} & \textbf{Notations} -- \textbf{Descriptions}    \\
			\hline
			
			\makecell[l]{Calligraphic fonts}  & 
			\makecell[l]{$\mathcal{D}$ -- datasets, $\mathcal{X}$ -- representations, \\ $\mathcal{T}$ -- the set of iTrees, $\mathcal{P}$ -- data pools in iTree nodes} \\
			
			Script typeface & 
			\makecell[l]{ 
				$\mathscr{F}$ -- anomaly scoring function,\\ $\mathscr{G}$ -- representation function} \\
			
			\makecell[l]{Bold lowercase letters} & 
			\makecell[l]{
				$\bm{o}$ -- original data objects, $\bm{x}$ -- representation vectors, \\ $\bm{p},\bm{q}$ -- random vectors used in CERE} \\ 
			\makecell[l]{Bold uppercase letters} & 
			\makecell[l]{ $\mathbf{W}$ -- weight matrices of neural networks, \\ $\mathbf{P},\mathbf{Q}$ -- matrices used in CERE } \\
			\makecell[l]{Others} & 
			\makecell[l]{$r$ -- number of representations, \\ 
				$t$ -- number of iTrees per representation, \\ 
				$T$ -- total number of iTrees,  \\
				$p(\cdot|\cdot)$ -- traversed node path,\\ 
				$g(\cdot|\cdot)$ -- averaged deviation degree
			} \\
			\hline
	\end{tabular}}%
	\label{tab:notation}%
\end{table}%

\section{Preliminaries: Isolation Forest}

For the sake of clarity, we recall the basic procedure of isolation forest (iForest) \cite{liu2008isolation}. 
A basic structure named isolation tree (iTree for short) is proposed. iTree $\tau$ is essentially a binary tree, and each node in the tree corresponds to a pool of data objects. 
A subset containing $n$ data objects is used as the data pool of the root node, which is randomly subsampled from the whole dataset. 
iTree $\tau$ grows by recursively isolating data objects in the leaf node (i.e., a disjoint partition of data objects into two child nodes) in a top-down fashion until remaining one data object in the node or reaching the maximum depth limit. 
iForest uses a simple isolation method that performs a comparison between the $j$-th dimension of the data object $\bm{o}^{(j)}$ and a splitting value $\eta$ as the branching criterion of each data object $\bm{o}$, where $j$ and $\eta$ respectively denote a randomly selected feature index and a split value within the range of available values of the $j$-th feature. 
Each data object $\bm{o}$ has a traversing path $p(\bm{o} | \tau)$ in the iTree $\tau$, and length of the traversed path, $ | p(\bm{o} | \tau) |$, can be naturally viewed as an indication of the abnormal degree of $\bm{o}$ (anomalies are often easier to be isolated with shorter path length). iForest constructs a forest of $T$ iTrees $\mathcal{T}$=$\{\tau_i\}_{i=1}^{T}$. The anomaly score of data object $\bm{o}$ is calculated based on its averaged path length $\mathbb{E}_{\tau_i \in \mathcal{T}}( |p(\bm{o} | \tau_i) |)$ over all of the iTrees in the forest $\mathcal{T}$, i.e., $\mathscr{F}_{\text{iFoerst}} (\bm{o} | \mathcal{T})\! =\! 2^{-\mathbb{E}_{\tau_i \in \mathcal{T}} \frac{ |p(\bm{o} | \tau_i)| }{C(T)}}$, where $C(T)$ is a normalising factor.

iForest uses a linear axis-parallel isolation method that only considers one dimension each time, and existing extension work introduces hyper-plane-based isolation that involves multiple dimensions, but similarly, only \textit{linear partition is admitted}.
These current isolation methods are limited to effectively handle hard anomalies that cannot be isolated using linear partitions on individual features or simple combinations of multiple features. 
Additionally, these existing methods generally suffer from the algorithmic bias brought by constraints hidden in their isolation strategies.

\section{\textsc{Deep Isolation Forest}}

We require a new isolation method that is unleashed from the above constraints so that it can effectively isolate those hard anomalies and avoid algorithmic bias. 
To this end, we introduce the \textsc{Deep Isolation Forest} (DIF) method. 
In a nutshell, DIF constructs an ensemble of representations derived from deep neural networks, and simple axis-parallel isolation is operated upon new data spaces to build iTrees in the forest. Instead of following the deep ensemble framework that combines independently trained neural networks, we use an ensemble of random representations produced by optimisation-free neural networks that only require simple casual initialisation.
This new representation scheme allows high freedom of the partition in the original data space. Each feature in the newly projected representation space is based on non-linear interactions among a number of original features, meaning that axis-parallel partitions of new space can form effective non-linear cuts in the original space. This way successfully liberates the slicing cuts from current linear constraints. Those hard anomalies that cannot be easily isolated in the original data space are possible to be exposed in representation spaces, and they can be isolated via fewer cuts and lead to clearer abnormality. 
Meanwhile, a unique synergy between random representations and random partition-based isolation can facilitate the overall ensemble-based abnormality estimation.

\subsection{Formulation of DIF}

DIF first produces the random representation ensemble via optimisation-free neural networks, which is defined as
\begin{equation}\label{eqn:rep}
	\mathscr{G}(\mathcal{D}) = \big\{ \mathcal{X}_u \subset \mathbb{R}^d \big \vert  \mathcal{X}_u = \phi_u(\mathcal{D};\theta_u)  \big\}_{u=1}^{r},
\end{equation}
where $r$ is the ensemble size, $\phi_u:\mathcal{D}\!\mapsto \!\mathbb{R}^d$ is the network that maps original data into new $d$-dimensional spaces, and the network weights in $\theta_u$ are randomly initialised.
Each representation is assigned with $t$ iTrees, and a forest $\mathcal{T}$=$\{ \tau_i \}_{i=1}^{T}$ containing $T$=$r$$\times$$t$ iTrees is constructed. 
iTree $\tau_i$ of $\mathcal{X}$ is initialised by a root node with a set of projected data $\mathcal{P}_{1}$ $\subset$ $\mathcal{X}$.
The $k$-th node with the data pool $\mathcal{P}_{k}$ is branched into two leaf nodes with disjoint subsets, i.e., $\mathcal{P}_{2k} \!=\! \{\bm{x} | \bm{x}^{(j_k)} \leq \eta_k,   \bm{x} \in \mathcal{P}_{k}  \}$ and $\mathcal{P}_{2k+1} \!=\! \{\bm{x} | \bm{x}^{(j_k)} \!>\! \eta_k,  \bm{x} \in \mathcal{P}_{k}   \}$, 
where $j_k$ is selected uniformly at random among all the dimensions of the newly created data space $\{1, \cdots, d\}$, $\bm{x}^{(j_k)}$ is the $j_k$-th dimension of the projected data object, and $\eta_k$ is a split value within the range 
of $\{ \bm{x}^{(j_k)} \vert  \bm{x} \in \mathcal{P}_{k}  \}$. 

After constructing $\mathcal{T}$, the abnormality of a data object $\bm{o}$ is evaluated by the isolation difficulty in each iTree of the forest $\mathcal{T}$. The scoring function is defined as
\begin{equation}\label{eqn:score_func_class}
	\mathscr{F}(\bm{o} | \mathcal{T}) = \Omega_{\tau_i \sim \mathcal{T}} I(\bm{o} | \tau_i),
\end{equation}
where $I(\bm{o} | \tau_i)$ denotes a function to measure the isolation difficulty in iTree $\tau_i$, and $\Omega$ denotes an integration function.

\subsection{Implementation of DIF}

There are two main components in DIF, i.e., random representation ensemble function $\mathscr{G}$ and isolation-based anomaly scoring function $\mathscr{F}$. 
To improve the time efficiency of representation function $\mathscr{G}$, we propose Computation-Efficient deep Representation Ensemble method (CERE), in which all the ensemble members can be computed simultaneously in a given mini-batch.
To further improve the accuracy of anomaly scoring, we propose Deviation-Enhanced Anomaly Scoring function (DEAS) by leveraging the hidden quantitative information enclosed in the projected dense representations along with qualitative comparisons.

\subsubsection{CERE: Computation-efficient Deep Representation Ensemble Method}

Successively feeding raw data into $r$ independent networks in Eq. (\ref{eqn:rep}) can induce a considerably high memory and time overhead. To inherit outstanding scalability of the original iForest, we introduce CERE to efficiently implement the representation ensemble function $\mathscr{G}(\mathcal{D})$.

Let $\mathbf{W}\in \mathbb{R}^{m \times n}$ be a weight matrix of a neural network layer, then following \cite{wen2019batchensemble}, we use a tuple of small random vectors $\bm{p}_i \in \mathbb{R}^m$ and $\bm{q}_i \in \mathbb{R}^n$ to yield a rank-one matrix via multiplication, which is used to derive the full weight matrices of each ensemble member. 
Formally, based on a base weight matrix $\mathbf{W}_0$, the weight matrix $\mathbf{W}_i$ of the $i$-th ensemble member is generated as
\begin{equation}
	\mathbf{W}_i = \mathbf{W}_0 \circ (\bm{p}_i \bm{q}_i^\top ) ,
\end{equation}
where $\circ$ denotes the Hadamard product.

The mapping process of incoming neurons $\bm{x}\in \mathbb{R}^{m}$ and the weight $\mathbf{W}_i$ can be further derived as follows:
\begin{equation}
	\begin{split}
		\mathbf{W}_i^{\top} \bm{x}   &= (\mathbf{W}_0 \circ \bm{p}_i \bm{q}_i^\top)^{\top} \bm{x} \\
		&= \mathbf{W}_0^{\top} (\bm{x} \circ \bm{p}_i) \circ \bm{q}_i .
	\end{split}
\end{equation}

Given $r$ tuples of weight vectors $\{ \langle \bm{p}_i , \bm{q}_i\rangle\}_{i=1}^{r}$ and a mini-batch of data $\mathcal{X}\in \mathbb{R}^{b \times m}$ with mini-batch size $b$ and dimension $m$, the ensemble of mapped results $\{ \mathcal{X}\mathbf{W}_1, \cdots, \mathcal{X}\mathbf{W}_r \}$ can be calculated via
\begin{equation}\label{eqn:batch}
	\begin{split}
		\begin{bmatrix}
			\mathcal{X}\mathbf{W}_1 \\
			\mathcal{X}\mathbf{W}_2 \\
			\cdots \\
			\mathcal{X} \mathbf{W}_r \\
		\end{bmatrix} 
		=
		\Bigg ( \bigg ( 
		\begin{bmatrix}
			\mathcal{X} \\
			\mathcal{X} \\
			\cdots \\
			\mathcal{X} \\
		\end{bmatrix} 
		\circ
		\begin{bmatrix}
			\mathbf{P}_1 \\
			\mathbf{P}_2 \\
			\cdots \\
			\mathbf{P}_r \\
		\end{bmatrix}
		\bigg )
		\mathbf{W}_0 \Bigg ) 
		\circ
		\begin{bmatrix}
			\mathbf{Q}_1 \\
			\mathbf{Q}_2 \\
			\cdots \\
			\mathbf{Q}_r \\
		\end{bmatrix},
	\end{split}
\end{equation}
where each row in $\mathbf{P}_i$ and $\mathbf{Q}_i$ is duplicated $\bm{p}_i$ and $\bm{q}_i$. Let $\mathbf{1}_b$ be a all-one vector with size $b$, $\mathbf{P}_i$ and $\mathbf{Q}_i$ are obtained via $\mathbf{P}_i = \mathbf{1}_b \bm{p}_i^{\top}$ and $\mathbf{Q}_i = \mathbf{1}_b \bm{q}_i^{\top}$.

The above vectorisation derivation allows the deep representation ensemble process in DIF to be efficiently calculated.
With the help of CERE, the time complexity of the ensemble process is similar to the feed-forward process of a single neural network since all the ensemble members can be computed simultaneously in a given mini-batch. 
Eq. (\ref{eqn:batch}) requires additional Hadamard product steps, but this operation is very cheap compared to matrix multiplication. 
One possible limitation is the batch size in Eq. (\ref{eqn:batch}). A mini-batch of $r \times t$ objects is simultaneously computed, which is larger than conventional settings. 
However, as computation within a mini-batch is automatically parallelisable, increasing the batch size incurs almost no time overhead. 
As for memory cost, this process is also feasible in a typical device because DIF does not involve optimisation, i.e., gradients are not calculated and saved. For example, a dataset with 10,000 features costs about 3GB of memory when using the recommended ensemble size $r$=$50$ and the batch size $b$=$64$.

Let $\Phi$ be the $L$-layer neural network using the newly-defined feed-forward step in Eq. (\ref{eqn:batch}). The ensemble of representations can be directly generated as 
\begin{equation}
	\mathscr{G}_{\text{CERE}}(\mathcal{D}) = \Phi \big( \mathcal{D} ; \Theta \big) = \big\{ \mathcal{X}_i \subset \mathbb{R}^d \big\}_{i=1}^{r} ,
\end{equation}
where $\Theta\! =\! \big\{ \mathbf{W}_l, \{\bm{p}_{(l,i)}\}_{i=1}^{r}, \{\bm{q}_{(l,i)}\}_{i=1}^{r} \big \}_{l=1}^{L}$. Note that other operations like activation or pooling and some layers that do not use a weight matrix are processed sequentially.

\subsubsection{DEAS: Deviation-enhanced Anomaly Scoring Function}

We further introduce a new anomaly scoring function DEAS. 
Recall that the standard anomaly scoring process in iForest only uses the length of the traversed path, i.e., all nodes are considered to have the same importance. The path length only provides limited information, which may not sufficiently delineate the isolation difficulty of data objects.
Except for qualitative comparison in each node, additional quantitative information is readily available to be leveraged, such as relation between the feature values of data objects and the branching threshold.

Motivated by this, we utilise the deviation degree of the feature value to the branching threshold as additional weighting information to further improve the measurement of isolation difficulty. These deviation degrees are important indicators to the isolation difficulty because the feature values in newly created data spaces are typically densely distributed and these deviations reflect the local density in the projected space., e.g., a small deviation value indicates that the slicing cut is on a dense region and thus is hard to isolate the data object.
Specifically, let $\bm{x}_u$ be the corresponding representation of a data object $\bm{o}$ in an iTree $\tau_i$. $p(\bm{x}_u | \tau_i) = \{1, \cdots, K\}$ is its traversed node path. 
We define the averaged deviation degree of $\bm{x}_u$ in $\tau_i$ as

\begin{equation}\label{eqn:deviation}
	g(\bm{x}_u | \tau_i)  = \frac{1}{ | p(\bm{x}_u | \tau_i) |} \sum_{k \in p(\bm{x}_u | \tau_i)} \vert \bm{x}_u^{(j_{k})} -  \eta_{k} \vert .
\end{equation}

We further combine the path length $|p(\bm{x}_u | \tau_i)|$ as in iForest and the deviation measure in Eq. (\ref{eqn:deviation}) to specify the function in Eq. (\ref{eqn:score_func_class}) by defining our deviation-enhanced isolation anomaly scoring function as:
\begin{equation}\label{eqn:scoring}
	\mathscr{F}_{\text{DEAS}}(\bm{o}|\mathcal{T}) 
	= 2^{
		-\mathbb{E}_{\tau_i \in \mathcal{T}} \frac{ |p(\bm{x}_u| \tau_i)|}{C(T)}
	} 
	\times 
	\mathbb{E}_{\tau_i \in \mathcal{T}}\big(g(\bm{x}_u | \tau_i) \big),
\end{equation}
where the first term is the averaged depth used as anomaly scores in iForest and the second term is the deviation-based anomaly score we introduce.

\subsection{Algorithm of DIF}
Algorithm \ref{alg1} presents the procedure of the construction of deep isolation trees $\mathcal{T}$.  Step (2) prepares $r$ random representations, and $t$ isolation trees $\{\tau_i\}_{i=1}^{t}$ are built upon each representation in Steps (4-15). 
For each isolation tree $\tau_i$, a subset of transformed data objects $\mathcal{P}_{1}$ is first randomly subsampled in Step (5) to initialise the root node. Each leaf node $P_k$ is then iteratively split by using a comparison branching criteria based on a randomly selected representation dimension $j_k$ and a split point $\eta_k$ in Steps (6-13).

We report the anomaly scoring procedure in Algorithm \ref{alg2}. 
Data object $\bm{o}$ is transformed to vectorised representations $\{\bm{x}\}_{u=1}^{r}$ in Step (1)
After the initialisation in Step (4), the data object traverses each tree $\tau_i$ by the criteria of each node and reaches the final node, during which the traverse path $p(\bm{x}_u|\tau_i)$ and the accumulated difference $\beta$ are recorded in Steps (5-12). The path length $| p(\bm{x}_u|\tau_i) |$ and the deviation $g(\bm{x}_u |\tau_i)$ in iTree $\tau_i$ are calculated in Step (13). The anomaly score of $\bm{o}$ is calculated and returned in Steps (16-17).

\renewcommand{\algorithmicrequire}{\textbf{Input:}}
\renewcommand{\algorithmicensure}{\textbf{Output:}}
\begin{algorithm}
	\caption{\textit{Construction of Deep Isolation Trees}}
	\label{alg1}
	\begin{algorithmic}[1]
		\REQUIRE $\mathcal{D}$ - input dataset
		\ENSURE $\mathcal{T}$ - forest of deep isolation trees
		\STATE Initialise $\mathcal{T}\leftarrow \varnothing $ 
		\STATE Generate representations $\{\mathcal{X}_{u}\}_{u=1}^{r}$ via $\mathscr{G}_{\text{CERE}}$
		\FOR{ $u=1$ to $r$ }
		\FOR{ $i=1$ to $t$}
		\STATE Initialise an isolation tree $\tau_i$ by setting the root node using $\mathcal{P}_{1} \subseteq \mathcal{X}_u$, $|\mathcal{P}_{1}| = n$ 
		\WHILE{$\mathcal{P}_{k}$ is a leaf node of tree $\tau_i$}
		\IF{$|\mathcal{P}_{k}|>1$ and the depth is smaller than $J$}
		\STATE Randomly select a dimension $j_k \in \{1, \cdots, d\}$
		\STATE Randomly select a split point $\eta_k$ between the \textit{max} and \textit{min} values of dimension $j_k$ in $\mathcal{P}_{k}$
		\STATE  $\mathcal{P}_{2k} \leftarrow \{\bm{x} | \bm{x}^{(j_k)} \leq \eta_k,  \bm{x} \in \mathcal{P}_{k}\}$
		\STATE  $\mathcal{P}_{2k+1} \leftarrow \{\bm{x} | \bm{x}^{(j_k)} > \eta_k, \bm{x} \in \mathcal{P}_{k}\} $
		\ENDIF
		\ENDWHILE
		\STATE $\mathcal{T} \leftarrow \mathcal{T} \cup \tau_i$
		\ENDFOR
		\ENDFOR
		\RETURN $\mathcal{T}$
	\end{algorithmic}
\end{algorithm}

\begin{algorithm}
	\caption{ \textit{Deviation-enhanced Anomaly Scoring}
	}
	\label{alg2}
	\begin{algorithmic}[1]
		\REQUIRE $\bm{o}$ - data object, $\mathcal{T}$ - set of deep isolation trees
		\ENSURE anomaly score $\mathscr{F}_{\text{DEAS}}(\bm{o}|\mathcal{T} )$
		\STATE Generate representations $\{\bm{x}_u\}_{u=i}^{r}$ via $\mathscr{G}_{\text{CERE}}$
		\FOR{ $u=1$ to $r$ }
		\FOR{ $i=1$ to $t$}
		\STATE Initialise $k \leftarrow 1$, $\beta \leftarrow 0$, $p(\bm{x}_u|\tau_i) \leftarrow \varnothing $
		\WHILE{$|\mathcal{P}_{k}| >1$ and not reaching $J$}
		\IF{$\bm{x}_u^{(j_k)} \leq \eta_k $ }
		\STATE $k \leftarrow 2k$
		\ELSE
		\STATE $k \leftarrow 2k+1$
		\ENDIF
		\STATE $p(\bm{x}_u|\tau_i)\leftarrow p(\bm{x}_u|\tau_i) \cup k$, $\beta \leftarrow \beta + |\bm{x}_u^{(j_k)}- \eta_k|$
		\ENDWHILE
		\STATE $g(\bm{x}_u|\tau_i) \leftarrow \beta / |p(\bm{x}_u|\tau_i)|$
		\ENDFOR
		\ENDFOR
		\RETURN $\mathscr{F}_{\text{DEAS}}(\bm{o}|\mathcal{T})\! \leftarrow \! 2^{ -\mathbb{E}_{\tau_i \in \mathcal{T}} \frac{ |p(\bm{x}_u | \tau_i)| }{C(T)}
		}\!  \times\! \mathbb{E}_{\tau \in \mathcal{T}}\big(g(\bm{x}_u | \tau_i) \big)$
	\end{algorithmic}
\end{algorithm}

\subsection{Theoretical Analysis}

\subsubsection{Time Complexity Analysis}
We first analyse the time complexity for the production of the random representation ensemble via CERE (i.e., Step 2 in Algorithm \ref{alg1}). 
Let the input data $\mathcal{D}$ be a tabular dataset with size $N \times D$. Multi-layer perceptron network is used for $\phi$. 
The used network $\phi$ comprises $L$ layers, and the $l$-th layer is with $d_l$ hidden units, the representation dimension is $d$. 
We use CERE to implement the ensemble with $r$ members within each mini-batch, and thus the whole feed-forward computation induces $O(r\times N \times (Dd_1 + d_ld + \sum_{l=1}^{L-1}d_ld_{l+1}))$. Only feed-forward steps are required in DIF, and the number of hidden units and representation dimension is generally small. 
Thus, this process is linear w.r.t. both data size and dimensionality, which does not introduce much extra computational overhead than the original iForest. 
In terms of the subsequent process of the iTree construction, given the depth limit $J$, we have the maximum $2^{J-1}$ splits (Steps (8-11)) during the growth of each iTree. For a node with $n$ samples, each split takes $O(n)$ complexity in determining the maximum and minimum value of the selected dimension and the assigning process. The overall process induces $O(2^{J-1}\times n \times r \times t)$. $J$ and $n$ often use fixed small values (typically 8 and 256 respectively). Therefore, the overall complexity is linear w.r.t. the ensemble size $r\times t$. As for the process in Algorithm \ref{alg2}, the traversing process takes a similar computation process, which has linear time complexity w.r.t. the size of testing sets and the ensemble size. 
Overall, according to the above analysis, the time complexity of DIF is $O(ND(r \times t))$. It has linear complexity w.r.t. data size, dimensionality, and ensemble size, which inherits desired scalability from iForest.

\subsubsection{DIF as a Generalisation of iForest and EIF}\label{sec:generalisation_analsis}

EIF \cite{hariri2019eif} is a recent extension of iForest \cite{liu2008isolation}, which has been shown to be a generalisation of iForest in \cite{hariri2019eif}. We show that DIF can be viewed as a further higher-level generalisation of isolation methods used in both iForest and EIF, i.e., the branching criteria used in iForest and EIF can be transformed into the format of DIF.

	Let $\bm{o} \in \mathbb{R}^{D}$ be a vectorised data object. Recall that the branching criterion in DIF is $\phi(\bm{o})^{(j)} \leq \eta$.
	Both iForest and EIF are special cases of DIF when the neural network $\phi$ is with one linear layer parameterised by a weight matrix $\mathbf{W}$, i.e., $\phi(\bm{o}) = \mathbf{W}^{\top}\bm{o}$.
	iForest splits the node by using the criterion $\bm{o}^{(j)} \leq \eta$, while DIF degrades to iForest if the weight matrix is set as an identity matrix, i.e., $\mathbf{W} = I_D$. 
	EIF uses a slicing hyper-plane for each node branching, and the slope of its hyper-plane is a normal vector $\bm{k} \in \mathbb{R}^D$, and $\bm{k}^{(i)} \sim \mathcal{N}(0, 1), \forall i \in \{1, \cdots D\}$. The intercept of the hyper-plane $\bm{p} \in \mathbb{R}^D$ is uniformly selected over the range of possible values at each branching point. The branching criterion is $(\bm{o} - \bm{p})\cdot \bm{k} \leq 0$, which is equivalent to $\bm{o}\cdot\bm{k} \leq \bm{p}\cdot\bm{k}$.
	We can fulfil exactly the same operation in DIF when the weight matrix satisfies $\mathbf{W} \in\mathbb{R}^{D\times 1}$. 
	The elements in $\mathbf{W}$ should also be initialised by a normal distribution $\mathcal{N}(0,1)$ to satisfy $\mathbf{W} = \bm{k}$. Additionally, the splitting point $\eta=\bm{p} \cdot \bm{k}$ can be understood as a standard random vector $\bm{p}$ sampled in possible values with a Gaussian noise $\bm{k}$.

\subsection{Discussions}

The power of DIF mainly depends on (i) the strong representation ability of neural networks, (ii) the discard of optimised representations, and (iii) the synergy between random representations and random partition-based isolation, which are respectively discussed below.

\subsubsection{Representation Ability of Neural Networks} 
\begin{figure}[t]
	\centering
	\includegraphics[width=0.49\textwidth]{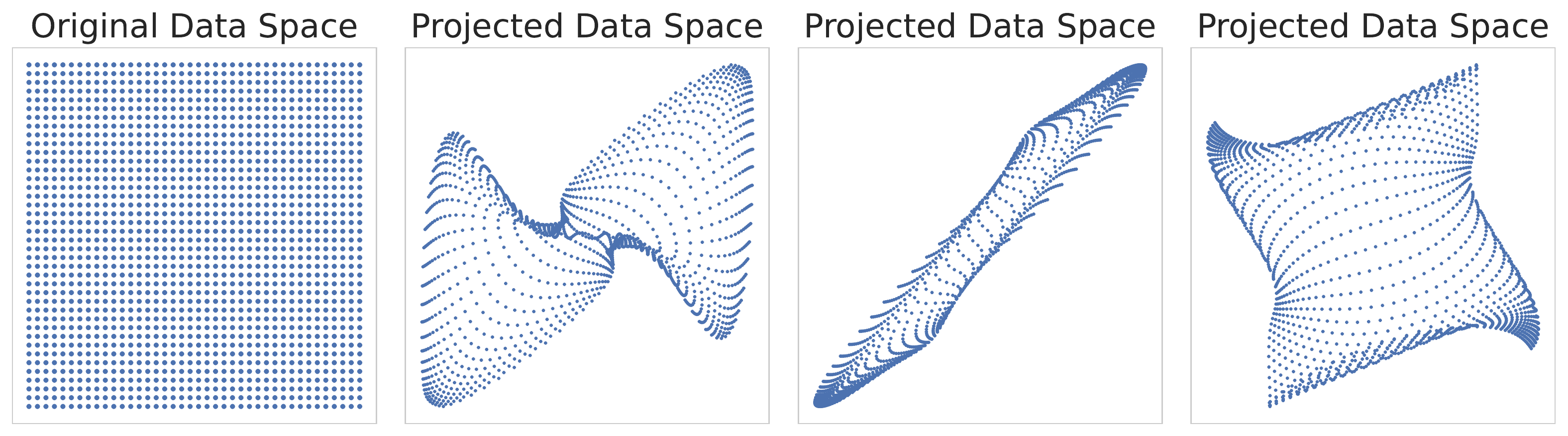}
	\caption{
		Data space transformation produced by casually initialised neural networks (original data space vs. three transferred data spaces). 
	}
	\label{fig:proj}
\end{figure}

Neural networks have strong representation power, even for randomly initialised networks. As shown in Fig. \ref{fig:proj} where the new data spaces are generated by random neural networks, the randomness in these initialised networks can create highly diversified data spaces, on which simple axis-parallel cuts can be equivalent to sophisticated slicing cuts in the original data space. Non-linear activation functions can effectively tweak and fold partition bounds to embed non-linearity into the isolation process, even though the networks are not optimised at all. 
On the other hand, there have been different deep learning architectures developed for various data types, so DIF is empowered to handle diverse data types by plugging the data-specific network backbone (e.g., multi-perceptron networks, recurrent networks, or graph neural networks) to produce the representations (see Sec. \ref{subsec: settings} for different neural networks used in DIF and Sec. \ref{sec:effectiveness} for their performance in different data types).

\subsubsection{Optimised vs. Casually Initialised Representations}
In general, we can use many representation learning networks specifically designed for anomaly detection, such as those in \cite{pang2018Learning,ruff2018dsvdd,wang2021rdp,xu2021beyond}, to obtain well-optimised feature representations. 
These representations are more expressive than randomly initialised representations if the optimisation objective well fits the input data.
However, instead of using optimised representations, DIF uses the casually initialised representations due to the following two main reasons.
(i) These loss functions are not versatile. It is difficult to devise one representation learning loss that can fit different data with diversified characteristics.
(ii) The downstream data partition might be strongly controlled by the optimisation process.
This way weakens the randomness and diversity of the feature representations, which are required in isolation-based anomaly scoring methods. These two intuitions are empirically investigated in our experiments by showing the performance of DIF on optimised representations produced by recent normality feature learning algorithms and the quality of these optimised representations  (see Sec. \ref{subsec:rep_scheme})

\subsubsection{Synergy between Random Representations and Random Partition-based Isolation }

DIF employs a novel representation scheme, i.e., the random representation ensemble produced via optimisation-free neural networks.
The parameters of these networks can be initialised by randomly sampling from widely-used initialisation distributions (e.g., normal or uniform distribution), easily yielding a set of feature representations with excellent randomness and diversity. Given a sufficiently large set of such random representations, we can largely boost the isolation power in the random data partition, making it possible to effectively isolate some really hard anomalies on some subsets of these representations. For example, as shown in Fig. \ref{fig:hardanom}, among a large set of new representation spaces, there are some selective new spaces where hard anomalies become easy-to-isolate data objects. Recall that isolation methods, including DIF, are based on an average measure for anomaly scoring. Thus, the anomalies would stand out in the anomaly scores as long as they are effectively isolated in some of the isolation trees.
DIF utilises this unique synergy between random representations and random partition-based isolation to largely improve the isolation process and subsequently the anomaly scoring function, 
resulting in significantly improved effectiveness of isolation-based anomaly detection.
We empirically investigate the significance of this synergy by respectively replacing random representations and random partition-based isolation with multiple alternatives (see Sec. \ref{sec:significance}).

\section{Experiments}
We now introduce our experimental analysis. This section is organised as follows:
In Sec. \ref{sec:setup}, we first start with the experimental setup including the used datasets, competing methods, parameter settings, and evaluation metrics;
Sec. \ref{sec:effectiveness}, \ref{sec:time}, and \ref{sec:robustness} evaluate the performance of our method w.r.t. effectiveness, scalability, and robustness; and in Sec. \ref{sec:significance} and \ref{sec:ablation}, we empirically analyse our method by investigating the significance of the synergy between random representations and random partition-based isolation and the contribution of CERE and DEAS.

\subsection{Experimental Setup}\label{sec:setup}
\subsubsection{Datasets}

We employ a large collection of publicly available and commonly-used real-world datasets, including ten tabular datasets, four graph datasets, and four time-series datasets. Their basic information is shown in Table \ref{tab:datainfo}.

\textbf{Tabular Data}.
\textit{Analysis}, \textit{Backdoor}, \textit{DoS}, and \textit{Exploits} are taken from a popular intrusion detection benchmark UNSW\_NB 15. Following \cite{pang2021toward,pang2019deep}, we select different attacks as anomalies in these datasets against normal network traffic. 
\textit{R8} is a highly-imbalanced text classification dataset, where the rare class are treated as anomalies by following \cite{pang2018Learning,rayana2016less}.
\textit{Cover} is from the ecology domain. \textit{Fraud} is for fraudulent credit card transaction detection. 
\textit{Pageblocks} and \textit{Shuttle} are provided by an anomaly benchmark study \cite{campos2016evaluation}.
\textit{Thrombin} is to detect unusual molecular bio-activity for drug design, which is an ultrahigh-dimensional anomaly detection dataset used in \cite{pang2018Learning}

\textbf{Graph Data}. We employ datasets from the popular graph benchmark Tox21, which is a project on toxicity evaluation of newly synthesised or used chemical compounds. These datasets are chosen since they are inherently imbalanced and contain real anomalies. The task is to detect abnormal graphs.

\textbf{Time-series Data}. Time series data are taken from the UCR time series anomaly archive \cite{wu2021current}.
We employ data with natural anomalies. 
\textit{Mars} is from NASA spacecraft. 
\textit{Gait} is sensor data of a subject who has Huntington’s disease (highly asymmetric gait), and anomalies are data from the weak leg.
\textit{ECG} is heartbeat data, where anomalies are ventricular beats, and \textit{ECG-wandering} (\textit{ECG-w} for short) is a long stretch of ECG with a wandering baseline.

\begin{table}[t]
	\centering
	\caption{Dataset information of the used tabular, graph, and time series (TS) datasets. $N$ indicates the number of data objects, with parenthesis denoting the pre-defined training/testing size of graph and time series datasets. $D$ denotes the number of dimensionality in tabular datasets, the average number of nodes per graph in graph datasets, and the number of sequences in each time series dataset. \#Anom (ratio) is the number of anomalies and the corresponding ratios. }
	\scalebox{0.9}{
		\begin{tabular}{lllll}
			\hline
			& \textbf{Data} & \textbf{$N$} & \textbf{$D$} & \textbf{\#Anom (ratio)} \\
			\hline
			\multirow{10}[0]{*}{\rotatebox{90}{Tabular}} 
			& Analysis & 95,677 & 197   & 2,677 (2.80\%) \\
			& Backdoor & 95,329 & 197   & 2,329 (2.44\%) \\
			& DoS   & 96,000 & 197   & 3,000 (3.13\%) \\
			& Exploits & 96,000 & 197   & 3,000 (3.13\%) \\
			& R8    & 3,974 & 9,468 & 51 (1.28\%) \\
			& Cover & 286,048 & 11    & 2,747 (0.96\%) \\
			& Fraud & 284,807 & 30    & 492 (0.17\%) \\
			& Pageblocks & 5,393 & 11    & 510 (9.46\%) \\
			& Shuttle & 1,013 & 10    & 13 (1.28\%) \\
			& Thrombin & 1,909 & 139,352 & 42 (2.20\%) \\
			\hline
			\multirow{4}[0]{*}{\rotatebox{90}{Graph}} & HSE   & 8,417 (8,150/267) & 17    & 10 (3.75\%) \\
			& MMP   & 7,558 (7,320/238) & 18    & 38 (0.50\%) \\
			& p53   & 8,903 (8,634/269) & 18    & 28 (10.41\%) \\
			& PPAR  & 8,451 (8,184/267) & 17    & 15 (5.62\%) \\
			\hline
			\multirow{4}[0]{*}{\rotatebox{90}{TS}} & Mars  & 11,349 (3,500/7,849) & 5     & 64 (0.82\%) \\
			& Gait  & 65,000 (22,167/42,833) & 3     & 401 (0.94\%) \\
			& Heart & 55,374 (20,185/35,189) & 3     & 201 (0.57\%) \\
			& Heart-w & 80,000 (20,000/60,000) & 1     & 251 (0.42\%) \\
			\hline
	\end{tabular}}%
	\label{tab:datainfo}%
\end{table}%

\subsubsection{Competing Methods}

To have a comprehensive comparison, DIF is compared with the following two types of anomaly detection approaches.

\textbf{iForest and Its Extensions (IF-based Methods)}. Apart from the popular iForest algorithm \cite{liu2008isolation}, its three advanced variants, i.e., EIF \cite{hariri2019eif}, PID \cite{gopalan2019pid}, and LeSiNN \cite{pang2015lesinn}, are employed. 
EIF slices data by using hyperplanes with random slopes and intercepts. In PID, the choice of splits is optimised based on the variance in the sparsity, and the anomaly scoring is redefined according to the sparsity. LeSiNN employs the nearest neighbour distance-based isolation ensemble, which is also known as aNNE in \cite{ting2017defying}.

\textbf{Ensemble of Deep Anomaly Detectors}.
\footnote{The ensemble performance of these deep models generally outperforms their solitary versions, and thus we focus on the comparison between ensemble-based results in the following experiments.}.
Different state-of-the-art (SOTA) deep anomaly detection methods are employed as base models, and we use the deep ensemble framework \cite{lakshminarayanan2017de} to construct a suite of ensemble-based deep contenders. 
For tabular data, we utilise four state-of-the-art deep methods including RDP \cite{wang2021rdp}, REPEN \cite{pang2018Learning}, Deep SVDD \cite{ruff2018dsvdd}, and a reconstruction-based Autoencoder baseline (RECON for short) \cite{aggarwal2017outlieranalysis}. 
We also employ the CERE ensemble method used in DIF for those methods to ensure their time efficiency and have a fair competition. 
As for graph data, a deep graph-level anomaly detector GLocalKD \cite{ma2022deep} is used. 
TranAD \cite{tuli2022tranad} is employed for time-series data.
Their ensemble versions are denoted as eRDP, eREPEN, eDSVDD, eRECON, eGLocalKD, and eTranAD. 
All these methods are specifically designed for anomaly detection on the corresponding data type.

\subsubsection{Parameter Settings and Implementations}\label{subsec: settings}

DIF uses 50 representations ($r$=$50$) and 6 isolation trees per representation ($t$=$6$), with 256 as subsampling size ($n$=$256$) for each iTree. DIF processes tabular data by using fully-connected multi-layer-perceptron networks. 
All of the IF-based competing methods use 300 trees (the ensemble size is the same as DIF). The subsampling size is set as 256. We use the maximum extension level of EIF, i.e., the extension level is adaptively set as the dimensionality minus 1. For LeSiNN, the subsampling size is 8 by following \cite{pang2015lesinn}. 
As IF-based competitors cannot directly process non-tabular data, we employ the latest powerful unsupervised representation learning models to yield high-quality vectorised representations, and specifically, InfoGraph \cite{sun2020infograph} and TS2Vec \cite{yue2022ts2vec} are respectively utilised for graph data and time series. 
Note that InfoGraph uses GIN as its graph encoder architecture, and TS2Vec applies a dilated CNN module with residual blocks. They propose specific learning objectives to optimise generated representations.  
For the sake of fairness, DIF also respectively utilises the same GIN and CNN network structure to handle graph data and time series but without any optimisation steps. 
The deep anomaly detectors are trained by 50 epochs and 30 steps per epoch. REPEN, DSVDD, and RECON take an Adam optimiser with a 1e-3 learning rate and use 64 objects per mini-batch, and we empirically found that RDP can work significantly better when using 1e-4. 
The default/recommended settings are used for the other parameters of these competing methods.

All the anomaly detection algorithms in our experiments are implemented using Python, with iForest from \texttt{scikit-learn} package, EIF from \texttt{eif} package, and other methods from their authors' releases.
The implementation of our method is publicly available\footnote{Source code of DIF can be downloaded from \url{https://github.com/xuhongzuo/deep-iforest}}.

\begin{table*}[htbp]
	\centering
	\caption{AUC-ROC and AUC-PR performance (mean $\pm$ standard deviation) of DIF and IF-based competing methods on ten real-world tabular datasets. PID and EIF runs out of memory (OOM) on the ultrahigh-dimensional dataset \textit{Thrombin}. The best performer is boldfaced. 
	}
	\scalebox{0.9}{
		\begin{tabular}{
				p{1.3cm} |
				p{1.4cm}<{\centering}p{1.3cm}<{\centering}
				p{1.3cm}<{\centering}p{1.3cm}<{\centering}
				p{1.3cm}<{\centering} |
				p{1.4cm}<{\centering}p{1.3cm}<{\centering}
				p{1.3cm}<{\centering}p{1.3cm}<{\centering}
				p{1.3cm}<{\centering} 
			}
			
			\hline
			\multirow{2}[0]{*}{\textbf{Data}} & \multicolumn{5}{c|}{\textbf{AUC-ROC}} & \multicolumn{5}{c}{\textbf{AUC-PR}} \\ 
			
			\cline{2-11}
			& \textbf{DIF (ours)} & \textbf{EIF} & \textbf{PID} & \textbf{LeSiNN} & \textbf{IF} & \textbf{DIF (ours)} & \textbf{EIF} & \textbf{PID} & \textbf{LeSiNN} & \textbf{IF} \\
			\hline
			
			Analysis & \textbf{0.931$_{\pm0.006}$} & 0.910$_{\pm0.005}$ & 0.820$_{\pm0.019}$ & 0.903$_{\pm0.008}$ & 0.782$_{\pm0.017}$ & \textbf{0.404$_{\pm0.051}$} & 0.198$_{\pm0.022}$ & 0.075$_{\pm0.007}$ & 0.183$_{\pm0.028}$ & 0.063$_{\pm0.006}$ \\
			Backdoor & \textbf{0.918$_{\pm0.002}$} & 0.902$_{\pm0.005}$ & 0.808$_{\pm0.016}$ & 0.894$_{\pm0.006}$ & 0.731$_{\pm0.021}$ & \textbf{0.453$_{\pm0.051}$} & 0.218$_{\pm0.028}$ & 0.066$_{\pm0.005}$ & 0.205$_{\pm0.031}$ & 0.046$_{\pm0.004}$ \\
			DoS   & \textbf{0.932$_{\pm0.003}$} & 0.918$_{\pm0.004}$ & 0.802$_{\pm0.013}$ & 0.896$_{\pm0.009}$ & 0.747$_{\pm0.020}$ & \textbf{0.440$_{\pm0.023}$} & 0.269$_{\pm0.027}$ & 0.075$_{\pm0.004}$ & 0.185$_{\pm0.028}$ & 0.060$_{\pm0.005}$ \\
			Exploits & \textbf{0.858$_{\pm0.010}$} & 0.840$_{\pm0.008}$ & 0.797$_{\pm0.011}$ & 0.816$_{\pm0.005}$ & 0.745$_{\pm0.010}$ & \textbf{0.273$_{\pm0.020}$} & 0.167$_{\pm0.011}$ & 0.077$_{\pm0.003}$ & 0.120$_{\pm0.013}$ & 0.062$_{\pm0.003}$ \\
			R8    & \textbf{0.930$_{\pm0.008}$} & 0.854$_{\pm0.006}$ & 0.881$_{\pm0.018}$ & 0.859$_{\pm0.001}$ & 0.853$_{\pm0.016}$ & \textbf{0.145$_{\pm0.031}$} & 0.101$_{\pm0.009}$ & 0.078$_{\pm0.011}$ & 0.094$_{\pm0.000}$ & 0.075$_{\pm0.008}$ \\
			Cover & \textbf{0.972$_{\pm0.010}$} & 0.872$_{\pm0.017}$ & 0.939$_{\pm0.007}$ & 0.885$_{\pm0.008}$ & 0.888$_{\pm0.017}$ & \textbf{0.246$_{\pm0.069}$} & 0.040$_{\pm0.006}$ & 0.069$_{\pm0.006}$ & 0.051$_{\pm0.004}$ & 0.055$_{\pm0.008}$ \\
			Fraud & \textbf{0.953$_{\pm0.002}$} & 0.950$_{\pm0.001}$ & 0.950$_{\pm0.002}$ & 0.952$_{\pm0.000}$ & 0.950$_{\pm0.001}$ & 0.387$_{\pm0.031}$ & 0.378$_{\pm0.027}$ & 0.186$_{\pm0.033}$ & \textbf{0.401$_{\pm0.001}$} & 0.155$_{\pm0.015}$ \\
			Pageblocks & \textbf{0.903$_{\pm0.006}$} & 0.902$_{\pm0.001}$ & 0.851$_{\pm0.003}$ & 0.887$_{\pm0.002}$ & 0.900$_{\pm0.005}$ & \textbf{0.547$_{\pm0.012}$} & 0.537$_{\pm0.006}$ & 0.421$_{\pm0.011}$ & 0.511$_{\pm0.007}$ & 0.476$_{\pm0.013}$ \\
			Shuttle & \textbf{0.941$_{\pm0.006}$} & 0.843$_{\pm0.009}$ & 0.864$_{\pm0.017}$ & 0.805$_{\pm0.005}$ & 0.862$_{\pm0.019}$ & \textbf{0.150$_{\pm0.017}$} & 0.061$_{\pm0.003}$ & 0.059$_{\pm0.008}$ & 0.048$_{\pm0.001}$ & 0.075$_{\pm0.014}$ \\
			Thrombin & \textbf{0.913$_{\pm0.003}$} & OOM   & OOM   & 0.912$_{\pm0.000}$ & 0.905$_{\pm0.002}$ & \textbf{0.468$_{\pm0.020}$} & OOM & OOM & 0.458$_{\pm0.001}$ & 0.372$_{\pm0.008}$ \\
			\hline
			\textit{Average}   & \textbf{0.925$_{\pm0.006}$} & 0.888$_{\pm0.006}$ & 0.857$_{\pm0.011}$ & 0.881$_{\pm0.004}$ & 0.836$_{\pm0.013}$ & \textbf{0.351$_{\pm0.033}$} & 0.219$_{\pm0.015}$ & 0.123$_{\pm0.010}$ & 0.226$_{\pm0.011}$ & 0.144$_{\pm0.008}$ \\
			\textit{p-value} & - & 0.004 & 0.004 & 0.002 & 0.002 &   -   & 0.004 & 0.004 & 0.006 & 0.002 \\
			\hline
			
	\end{tabular}}%
	\label{tab:tabular1}%
\end{table*}%

\begin{table*}[htbp]
	\centering
	\caption{AUC-ROC and AUC-PR performance of DIF and its deep ensemble-based competing methods. 
	}
	\scalebox{0.9}{
		\begin{tabular}{
				p{1.3cm} |
				p{1.4cm}<{\centering}p{1.3cm}<{\centering}
				p{1.3cm}<{\centering}p{1.3cm}<{\centering}
				p{1.3cm}<{\centering} |
				p{1.4cm}<{\centering}p{1.3cm}<{\centering}
				p{1.3cm}<{\centering}p{1.3cm}<{\centering}
				p{1.3cm}<{\centering} 
			}
			
			\hline
			\multirow{2}[0]{*}{\textbf{Data}} & \multicolumn{5}{c|}{\textbf{AUC-ROC}} & \multicolumn{5}{c}{\textbf{AUC-PR}} \\ 
			
			\cline{2-11}
			& \textbf{DIF (ours)} & \textbf{eRDP} & \textbf{eREPEN} & \textbf{eDSVDD} & \textbf{eRECON} &
			\textbf{DIF (ours)} & \textbf{eRDP} & \textbf{eREPEN} & \textbf{eDSVDD} & \textbf{eRECON}   \\
			\hline
			
			Analysis & \textbf{0.931$_{\pm0.006}$} & 0.899$_{\pm0.005}$ & 0.883$_{\pm0.026}$ & 0.844$_{\pm0.006}$ & 0.862$_{\pm0.002}$ & \textbf{0.404$_{\pm0.051}$} & 0.400$_{\pm0.019}$ & 0.168$_{\pm0.051}$ & 0.374$_{\pm0.028}$ & 0.093$_{\pm0.002}$ \\
			Backdoor & \textbf{0.918$_{\pm0.002}$} & 0.900$_{\pm0.006}$ & 0.863$_{\pm0.022}$ & 0.916$_{\pm0.004}$ & 0.840$_{\pm0.003}$ & \textbf{0.453$_{\pm0.051}$} & 0.422$_{\pm0.020}$ & 0.163$_{\pm0.055}$ & 0.448$_{\pm0.030}$ & 0.082$_{\pm0.004}$ \\
			DoS   & \textbf{0.932$_{\pm0.003}$} & 0.905$_{\pm0.004}$ & 0.861$_{\pm0.025}$ & 0.900$_{\pm0.004}$ & 0.856$_{\pm0.004}$ & \textbf{0.440$_{\pm0.023}$} & 0.420$_{\pm0.013}$ & 0.154$_{\pm0.047}$ & 0.409$_{\pm0.028}$ & 0.105$_{\pm0.004}$ \\
			Exploits & \textbf{0.858$_{\pm0.010}$} & 0.795$_{\pm0.008}$ & 0.744$_{\pm0.040}$ & 0.748$_{\pm0.017}$ & 0.808$_{\pm0.002}$ & \textbf{0.273$_{\pm0.020}$} & 0.250$_{\pm0.008}$ & 0.080$_{\pm0.019}$ & 0.249$_{\pm0.013}$ & 0.086$_{\pm0.001}$ \\
			R8    & \textbf{0.930$_{\pm0.008}$} & 0.864$_{\pm0.020}$ & 0.904$_{\pm0.004}$ & 0.883$_{\pm0.014}$ & 0.779$_{\pm0.000}$ & \textbf{0.145$_{\pm0.031}$} & 0.123$_{\pm0.017}$ & 0.086$_{\pm0.005}$ & 0.074$_{\pm0.006}$ & 0.082$_{\pm0.000}$ \\
			Cover & 0.972$_{\pm0.010}$ & 0.966$_{\pm0.008}$ & 0.896$_{\pm0.013}$ & \textbf{0.981$_{\pm0.005}$} & 0.922$_{\pm0.004}$ & 0.246$_{\pm0.069}$ & 0.192$_{\pm0.049}$ & 0.044$_{\pm0.006}$ & \textbf{0.353$_{\pm0.055}$} & 0.067$_{\pm0.003}$ \\
			Fraud & 0.953$_{\pm0.002}$ & 0.947$_{\pm0.002}$ & \textbf{0.954$_{\pm0.001}$} & 0.948$_{\pm0.002}$ & 0.946$_{\pm0.001}$ & 0.387$_{\pm0.039}$ & 0.329$_{\pm0.025}$ & 0.345$_{\pm0.030}$ & \textbf{0.480$_{\pm0.026}$} & 0.372$_{\pm0.002}$ \\
			Pageblocks & \textbf{0.903$_{\pm0.010}$} & 0.850$_{\pm0.005}$ & 0.896$_{\pm0.005}$ & 0.882$_{\pm0.005}$ & 0.829$_{\pm0.012}$ & \textbf{0.547$_{\pm0.020}$} & 0.466$_{\pm0.011}$ & 0.511$_{\pm0.019}$ & 0.397$_{\pm0.009}$ & 0.542$_{\pm0.032}$ \\
			Shuttle & \textbf{0.941$_{\pm0.006}$} & 0.900$_{\pm0.009}$ & 0.780$_{\pm0.025}$ & 0.837$_{\pm0.037}$ & 0.779$_{\pm0.014}$ & \textbf{0.150$_{\pm0.017}$} & 0.132$_{\pm0.009}$ & 0.031$_{\pm0.003}$ & 0.144$_{\pm0.022}$ & 0.044$_{\pm0.002}$ \\
			Thrombin & 0.913$_{\pm0.003}$ & 0.869$_{\pm0.019}$ & \textbf{0.914$_{\pm0.004}$} & 0.366$_{\pm0.019}$ & 0.911$_{\pm0.000}$ & \textbf{0.468$_{\pm0.020}$} & 0.368$_{\pm0.053}$ & 0.399$_{\pm0.056}$ & 0.016$_{\pm0.001}$ & 0.457$_{\pm0.000}$ \\
			\hline
			\textit{Average}   & \textbf{0.925$_{\pm0.006}$} & 0.889$_{\pm0.009}$ & 0.870$_{\pm0.016}$ & 0.831$_{\pm0.011}$ & 0.853$_{\pm0.004}$ & \textbf{0.351$_{\pm0.034}$} & 0.310$_{\pm0.022}$ & 0.198$_{\pm0.029}$ & 0.294$_{\pm0.022}$ & 0.193$_{\pm0.005}$ \\
			
			\textit{p-value} & - & 0.002 & 0.01  & 0.01  & 0.002 &    -   & 0.002 & 0.002 & 0.232 & 0.002 \\
			
			\hline
	\end{tabular}}%
	\label{tab:tabular2}%
\end{table*}%

\subsubsection{Evaluation Metrics and Computing Infrastructure}\label{sec:metric}
Following the mainstream evaluation protocols of anomaly detection \cite{pang2019deep,hariri2019eif,liu2008isolation}, the detection accuracy is evaluated by two complementary metrics including Area Under the ROC Curve (AUC-ROC) and Area Under the PR Curve (AUC-PR).
ROC curve indicates the true positives against false positives, while PR curve summarises precision and recall of the anomaly class only. 
The paired \textit{Wilcoxon} signed rank test is used to examine the statistical significance of the performance of DIF against each competing method.

We then introduce a new metric called Anomaly Isoability Index ($AII$) to measure the quality of representations.
We borrow the concept of triplet loss \cite{schroff2015triplet} to count the percentage of effectively isolated anomalies among all true anomalies in each representation space, i.e.,
\begin{equation}
	AII = 
	\mathop{\mathbf{P}}\limits_{\bm{a} \sim \mathcal{A}} \Big( \mathop{median}\limits_{\bm{n}_i \in \mathcal{N}} \big \{\frac{1}{|\mathcal{C}|} \sum_{\bm{n}_j \in \mathcal{C}} \big( \omega(\bm{a} | \bm{n}_i, \bm{n}_j) \big) \big \} >0 \Big ),
\end{equation}
where $\omega(\bm{a} | \bm{n}_i, \bm{n}_j)$ = $d(\bm{a}, \bm{n}_i)\! -\! d(\bm{n}_j, \bm{n}_j)$ denotes the difference between the Euclidean distances of the two pairs, $\bm{a}\!\sim\!\mathcal{A}$ represents any anomaly drawn from the true anomaly set in the dataset, 
$\mathcal{C}$ is a group of randomly sampled normal anchors, and $\mathcal{N}$ is another set of random normal samples to delegate the whole normal distribution. All the above data objects are from the target representation space.
$|\mathcal{C}|$=$20$ and $|\mathcal{N}|$=$1000$ are used as we found empirically that these two settings are sufficiently large to compute the $AII$ metric.

The computational time of all methods is based on a workstation with Intel Xeon Silver 4210R CPU, a single NVIDIA TITAN RTX GPU, and 64 GB RAM.

\begin{table*}[htbp]
	\centering
	\caption{Results on detecting abnormal graphs and anomalies in time series.}
	\scalebox{0.9}{
		\begin{tabular}{
				p{0.3cm} p{1.2cm} |
				p{1.4cm}<{\centering}p{1.3cm}<{\centering}
				p{1.3cm}<{\centering}p{1.3cm}<{\centering}
				p{1.3cm}<{\centering}|
				p{1.4cm}<{\centering}p{1.3cm}<{\centering}
				p{1.3cm}<{\centering}p{1.3cm}<{\centering}
				p{1.3cm}<{\centering}}
			
			\hline
			
			& \multirow{2}[0]{*}{\textbf{Data}}  & \multicolumn{5}{c|}{\textbf{AUC-ROC}}  & \multicolumn{5}{c}{\textbf{AUC-PR}} \\
			\cline{3-12}
			& & \textbf{DIF (ours)} & \textbf{EIF} & \textbf{LeSiNN} & \textbf{iForest} & \textbf{eGLocalKD} & \textbf{DIF (ours)} & \textbf{EIF} & \textbf{LeSiNN} & \textbf{iForest} & \textbf{eGLocalKD} \\
			\hline
			\multirow{4}[0]{*}{\rotatebox{90}{Graph}} 
			& HSE   & \textbf{0.737$_{\pm0.013}$} & 0.715$_{\pm0.014}$ & 0.702$_{\pm0.001}$ & 0.697$_{\pm0.014}$ & 0.593$_{\pm0.002}$ & \textbf{0.094$_{\pm0.005}$} & 0.088$_{\pm0.004}$ & 0.084$_{\pm0.000}$ & 0.082$_{\pm0.004}$ & 0.054$_{\pm0.000}$ \\
			& MMP   & \textbf{0.715$_{\pm0.006}$} & 0.663$_{\pm0.012}$ & 0.666$_{\pm0.000}$ & 0.667$_{\pm0.018}$ & 0.675$_{\pm0.001}$ & \textbf{0.260$_{\pm0.006}$} & 0.216$_{\pm0.006}$ & 0.217$_{\pm0.000}$ & 0.219$_{\pm0.011}$ & 0.233$_{\pm0.001}$ \\
			& p53   & \textbf{0.680$_{\pm0.008}$} & 0.597$_{\pm0.017}$ & 0.606$_{\pm0.000}$ & 0.619$_{\pm0.013}$ & 0.640$_{\pm0.001}$ & \textbf{0.177$_{\pm0.006}$} & 0.138$_{\pm0.004}$ & 0.144$_{\pm0.000}$ & 0.143$_{\pm0.004}$ & 0.150$_{\pm0.000}$ \\
			& PPAR  & 0.701$_{\pm0.013}$ & 0.716$_{\pm0.005}$ & 0.711$_{\pm0.000}$ & \textbf{0.733$_{\pm0.009}$} & 0.643$_{\pm0.001}$ & 0.127$_{\pm0.008}$ & 0.173$_{\pm0.006}$ & 0.165$_{\pm0.001}$ & \textbf{0.208$_{\pm0.012}$} & 0.086$_{\pm0.000}$ \\
			
			\midrule		
			&	& \textbf{DIF (ours)} & \textbf{EIF} & \textbf{LeSiNN} & \textbf{iForest} & \textbf{eTranAD} & \textbf{DIF (ours)} & \textbf{EIF} & \textbf{LeSiNN} & \textbf{iForest} & \textbf{eTranAD} \\
			\hline
			
			\multirow{4}[0]{*}{\rotatebox{90}{TS}} &	Mars  & 0.952$_{\pm0.017}$ & \textbf{0.980$_{\pm0.006}$} & 0.942$_{\pm0.014}$ & 0.947$_{\pm0.015}$ & 0.947$_{\pm0.016}$ & \textbf{0.626$_{\pm0.024}$} & 0.458$_{\pm0.031}$ & 0.400$_{\pm0.009}$ & 0.390$_{\pm0.043}$ & 0.334$_{\pm0.020}$ \\
			& Gait  & \textbf{0.998$_{\pm0.001}$} & 0.997$_{\pm0.001}$ & 0.998$_{\pm0.000}$ & 0.997$_{\pm0.001}$ & 0.998$_{\pm0.000}$ & \textbf{0.835$_{\pm0.064}$} & 0.772$_{\pm0.048}$ & 0.829$_{\pm0.010}$ & 0.741$_{\pm0.073}$ & 0.806$_{\pm0.010}$ \\
			& ECG & \textbf{0.997$_{\pm0.001}$} & 0.986$_{\pm0.001}$ & 0.987$_{\pm0.000}$ & 0.987$_{\pm0.001}$ & 0.976$_{\pm0.001}$ & \textbf{0.809$_{\pm0.031}$} & 0.705$_{\pm0.004}$ & 0.710$_{\pm0.000}$ & 0.711$_{\pm0.002}$ & 0.692$_{\pm0.001}$ \\
			& ECG-w & \textbf{1.000$_{\pm0.000}$} & 0.988$_{\pm0.001}$ & 0.985$_{\pm0.001}$ & 0.981$_{\pm0.001}$ & 0.990$_{\pm0.000}$ & \textbf{1.000$_{\pm0.000}$} & 0.255$_{\pm0.011}$ & 0.219$_{\pm0.008}$ & 0.181$_{\pm0.005}$ & 0.297$_{\pm0.001}$ \\
			\hline
	\end{tabular}}%
	\label{tab:graphts}%
\end{table*}%

\subsection{Effectiveness in Reducing False Negatives}\label{sec:effectiveness}

\subsubsection{Tabular Data}

Table \ref{tab:tabular1} and \ref{tab:tabular2} present the AUC-ROC and AUC-PR results of our method DIF and eight competing methods. 
Overall, DIF largely reduces the false negatives compared to iForest and its three extensions, resulting in substantial averaged performance improvement in detection precision and/or recall rates, and thus, DIF obtains superior AUC-PR and AUC-ROC performance. Particularly, in the average AUC-PR, DIF significantly outperforms EIF (61\%), PID (186\%), LeSiNN (56\%), iForest (144\%), eRDP (13\%), eREPEN (77\%), eDSVDD (19\%), and eRECON (82\%). DIF also obtains 4\% - 11\% AUC-ROC improvement across these methods.

As shown in Table \ref{tab:tabular1}, DIF significantly outperforms isolation-based methods at the 99\% confidence level.
DIF is the best isolation-based detector across all the ten datasets except AUC-PR on \textit{Fraud}, on which the nearest neighbour-based anomaly measure LeSiNN is more effective. The features in \textit{Fraud} are the results of PCA transformation due to the confidentiality issues, and thus the distance concept used in the nearest neighbour information can well reflect the proximity relationship.
By contrast, DIF does not rely on such prior information that is not always reliable in all the datasets. These results demonstrate superior isolation power of DIF, which can effectively isolate anomalies that may not be possible in existing IF-based methods due to the challenges like data sparsity and non-linearity. This is particularly true on challenging high-dimensional datasets like \textit{R8}, \textit{Analysis}, \textit{Backdoor} and \textit{DoS}.

Compared to deep ensemble-based methods, DIF performs significantly better at the 99\% confidence level according to the AUC-ROC performance in Table \ref{tab:tabular2}.
DIF achieves the best performance on seven out of ten datasets in terms of both AUC-ROC and AUC-PR, and it obtains very competitive results on the rest three datasets with less than 0.01 difference in AUC-ROC.
Nevertheless, the comparison results between DIF and its deep ensemble-based counterparts are very encouraging given the fact that DIF does not involve any optimisation while these deep methods need to be properly trained using pre-defined objective functions to better expose anomalies. 
The superiority of DIF in this comparison owes to the representation diversity of each ensemble member and the unique synergy between random representations and random partition-based isolation in the downstream anomaly scoring process. 
More importantly, DIF runs significantly faster than these deep ensemble-based contenders by around two orders of magnitude (see Sec. \ref{sec:time}).
Its superior computational efficiency endows DIF with stronger practicability in real-world applications.

\subsubsection{Graph Data and Time Series}

As reported in Sec. \ref{subsec: settings}, traditional isolation-based methods are performed upon vectorised representations learned by unsupervised representation learning method InfoGraph \cite{sun2020infograph} and TS2Vec \cite{yue2022ts2vec}.
DIF respectively utilises the same network structure (GIN and dilated CNN) in InfoGraph and TS2Vec but without any optimisation steps. 
GLocalKD \cite{ma2022deep} is a graph-level anomaly detector, and TranAD \cite{tuli2022tranad} is a deep anomaly detection method specifically designed for time series. We also employ the ensemble version of these two SOTA methods as competitors.

Detection performance on graph data and time series are shown in Table \ref{tab:graphts}, in which EIF and LeSiNN are selected as contenders due to their preferable performance in tabular datasets over PID, with iForest also included as a baseline. 
DIF is the best performer on three out of four datasets in both graph-level and time series anomaly detection tasks. This performance of DIF is remarkable in that it outperforms not only the isolation-based methods that are empowered by the latest powerful representation learning models but the recent SOTA methods that are specifically designed to extensively learn data-type-specific characteristics (e.g., holistic graph structure or temporal dependence) for effectively detecting these graph/sequential anomalies. By contrast, DIF performs well in a unified framework by only replacing the network structure with a different randomly initialised network backbone, offering a significantly simpler yet data-type-agnostic effective solution.

\begin{figure}[t]
	\centering
	\includegraphics[width=0.49\textwidth]{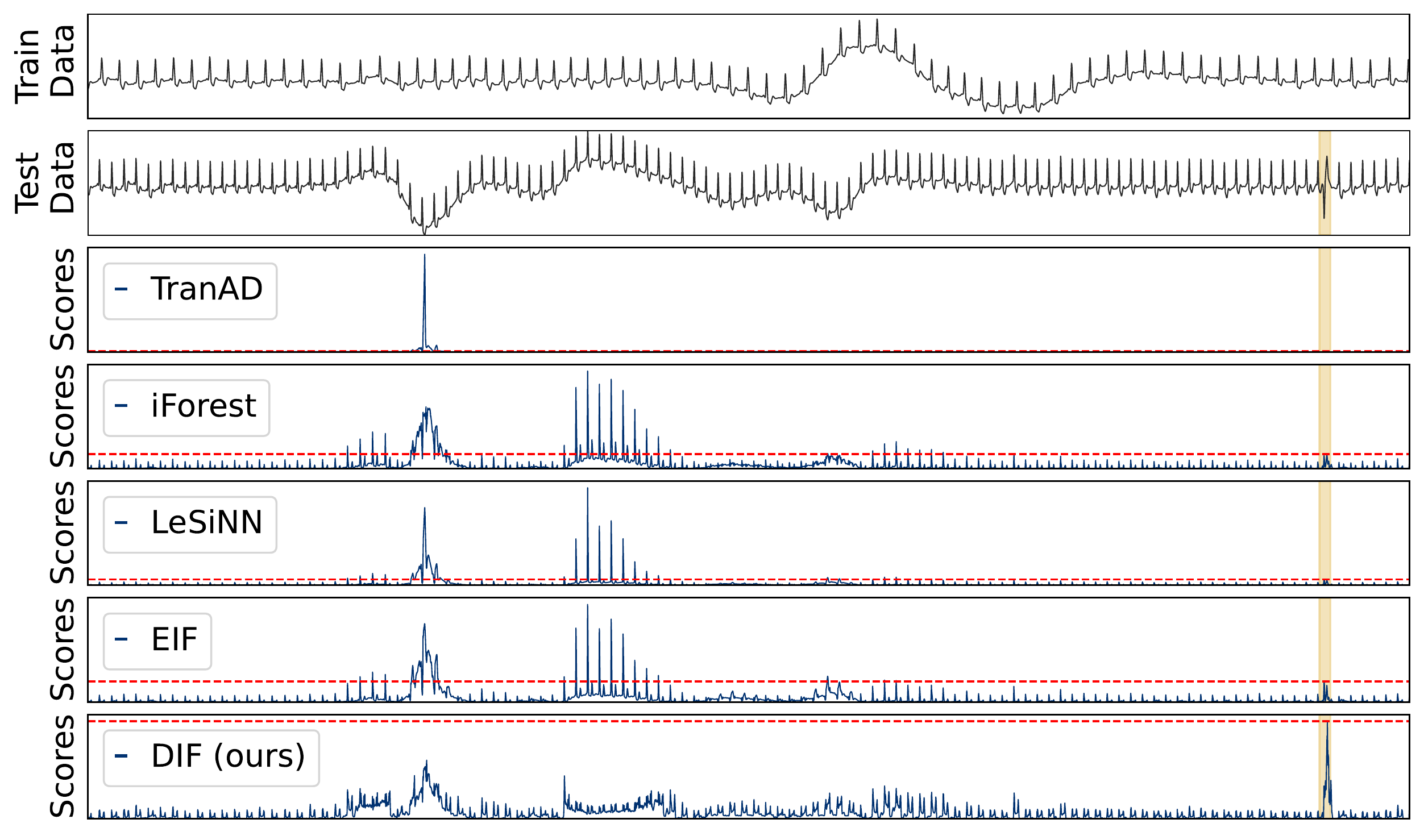}
	\caption{(\textbf{Top to bottom}) Snippets of training/testing data of \textit{ECG-w} with wandering baseline and detection results, i.e., anomaly scores, of DIF and its competitors. Anomalous heartbeats are highlighted in golden yellow. Red dashed lines indicate the reported highest anomaly scores in this anomalous duration. }
	\label{fig:case}
\end{figure}

We further show how DIF works by visualising its results on a time series dataset. Fig. \ref{fig:case} visualises the ECG-w data (the id is \texttt{sddb49} in the UCR benchmark) and the detection results of DIF and its competing methods. As illustrated in the introduction of the benchmark, this dataset is a challenging case that may confuse many algorithms due to its wandering shape. Albeit wandering baseline in testing data, we can also see this in the training set, and thus detection models are expected to tolerate these noisy regions. Our method DIF reduces the false negatives by successfully yielding significantly higher anomaly scores on anomalous heartbeats highlighted in golden yellow. In comparison, competing methods are misled by the fluctuating trends, showing much higher scores on noisy regions and omitting real anomalies.

\begin{figure}[t]
	\centering
	\includegraphics[width=0.49\textwidth]{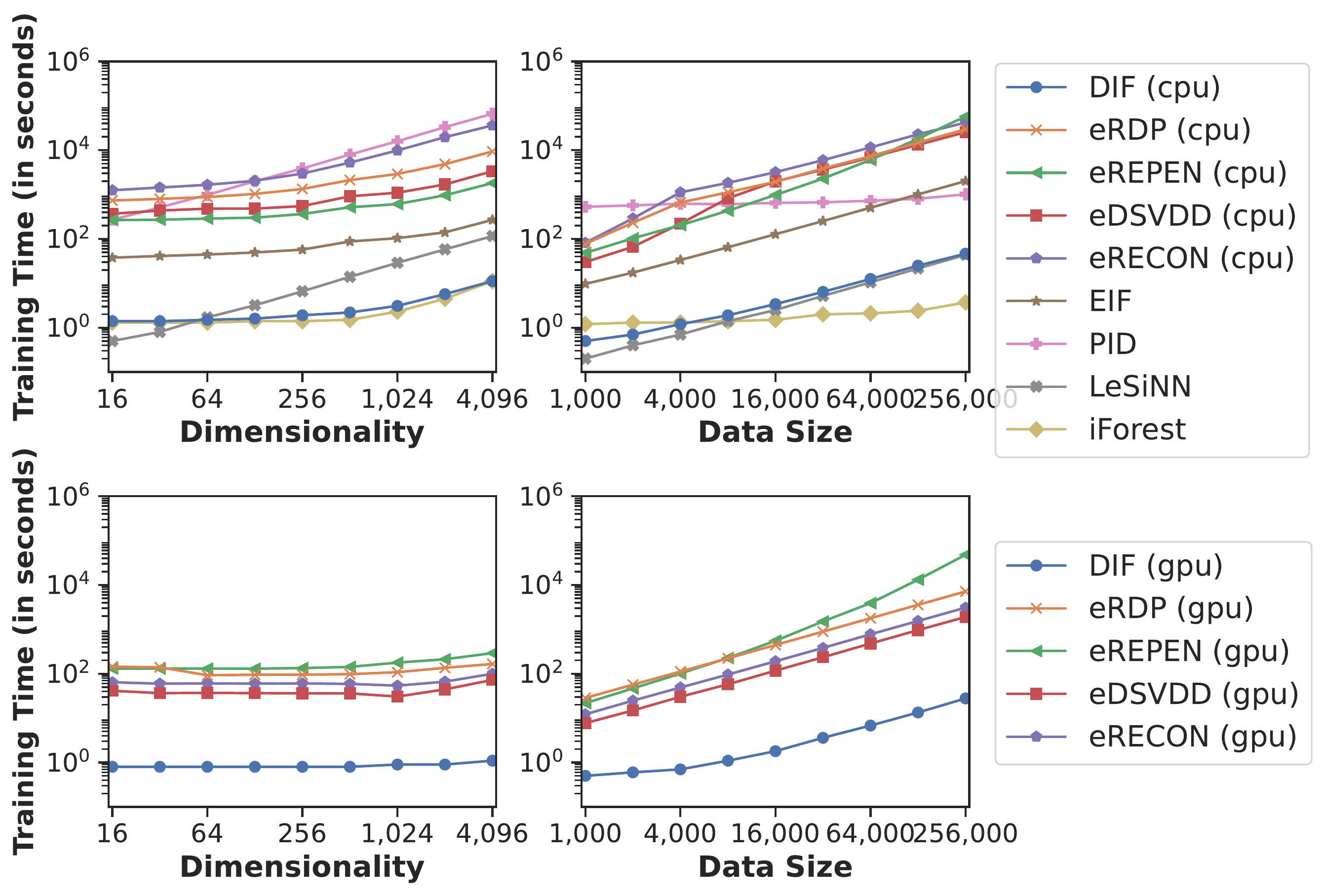}
	\caption{Scalability test results. (\textbf{Top}) The training time of all the anomaly detectors on a CPU device; and (\textbf{Bottom}) The results of deep ensemble-based methods (including DIF) on a GPU device. }
	\label{fig:scal}
\end{figure}

\subsection{Scalability to High-dimensional, Large-scale Data}\label{sec:time}

This experiment examines the scalability of DIF and its contenders.
These anomaly detectors are performed on a group of synthetic tabular datasets with different sizes and dimensionalities to record their training times. Nine datasets are with 5,000 data objects, and their dimensions range from 16 to 4,096. The other nine datasets are with 32 features and varied sizes from a minimum of 1,000 up to 256,000.
IF-based anomaly detectors only need CPU devices, while deep ensemble-based methods can leverage GPU acceleration. Therefore, we report the training time of DIF and its deep competitors on both GPU and CPU devices.

\begin{figure*}[htbp]
	\centering 
	\subfigcapskip=-5pt %
	\subfigure[DIF vs. IF-based Competing Methods]{
		\includegraphics[width=0.49\linewidth]{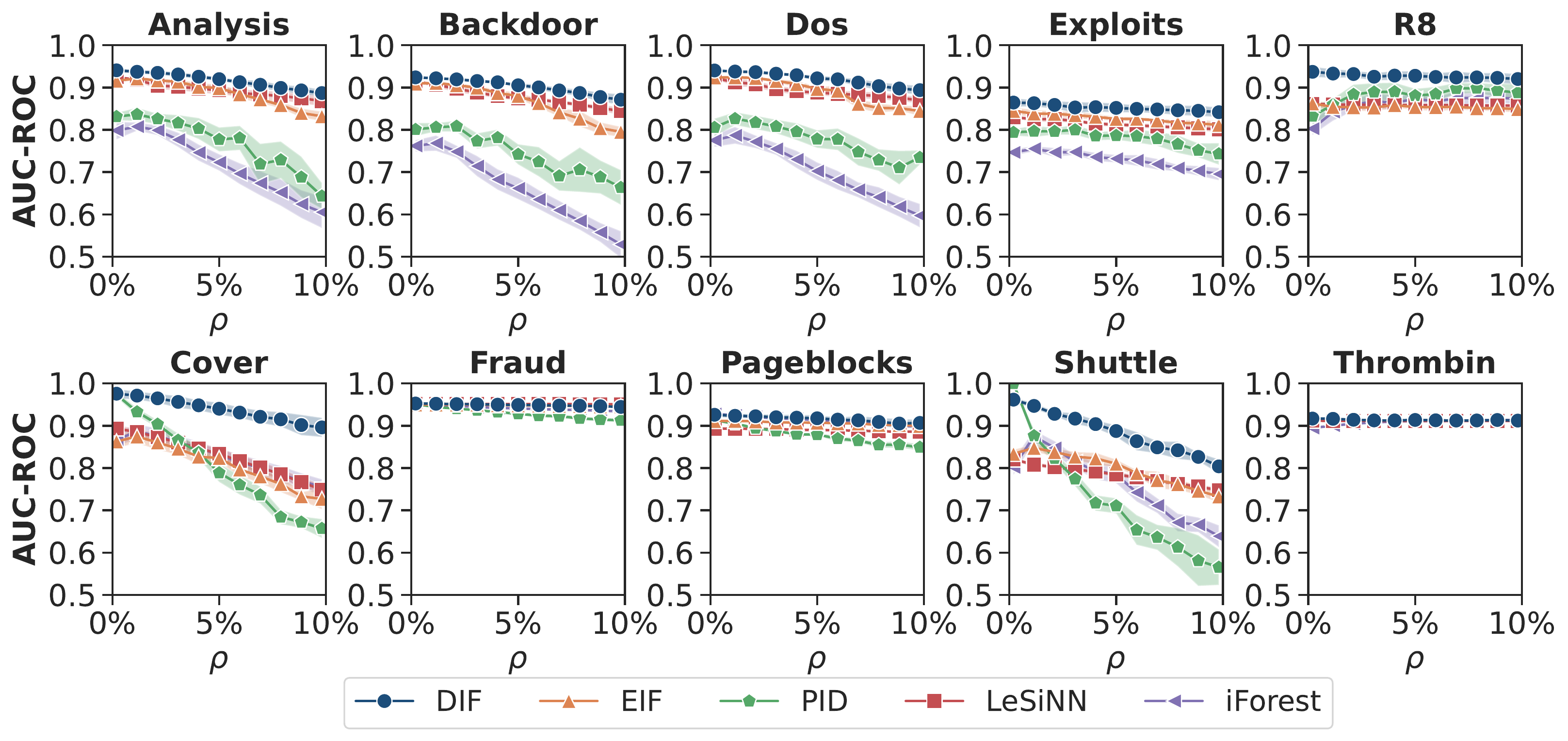}} 
	\subfigure[DIF vs. Deep Ensemble-based Competing Methods]{
		\includegraphics[width=0.49\linewidth]{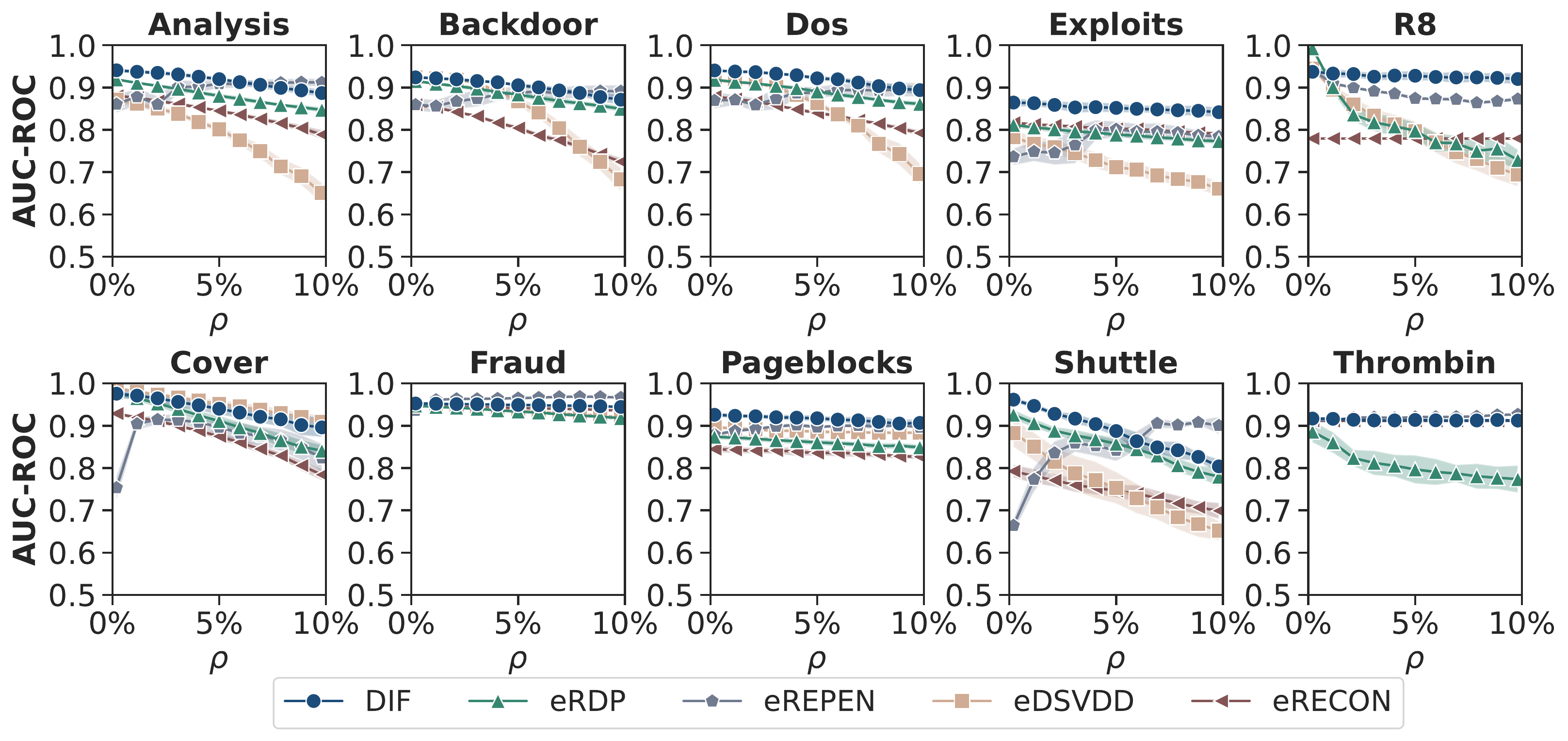}}
	\caption{AUC-ROC w.r.t. different contamination ratios $\rho$ (the percentage of anomalies in the training set). 
	}
	\label{fig:robustness}
\end{figure*}

Fig. \ref{fig:scal} (top) shows the scalability test results of DIF and its competing methods on a CPU device, and Fig. \ref{fig:scal} (bottom) reports the comparison using GPU. 
DIF and all the other isolation-based methods present good scalability w.r.t. both dimensionality and data size compared to deep ensemble-based methods when using CPU computation. This is owed to the subsampling-based methods applied to both data samples and dimensions in isolation-based detectors. 
Besides, deep ensemble-based methods can greatly benefit from GPU acceleration when handling high-dimensional data. DIF shows outstanding time efficiency compared to its deep ensemble-based counterparts since it only requires one feed-forward step instead of a large number of training epochs. 
These results demonstrate that DIF inherits excellent computational efficiency from the iForest. Note that DIF can obtain almost the same scalability as iForest w.r.t. the data dimensionality, as DIF performs isolation on newly projected spaces of much smaller dimensionality while iForest works on the original data space.

\subsection{Robustness w.r.t. Anomaly Contamination}\label{sec:robustness}
This experiment examines the performance of DIF and its contenders when datasets containing different anomaly contamination ratios. 
Time-series datasets and graph datasets have pre-defined train-test split and their training sets do not contain anomalies, and thus we use tabular datasets as our bases here.
Following \cite{pang2021toward,pang2019deep}, we adjust the contamination ratios by injecting/removing anomalies, such that ratios range from 0\% to 10\%.
These detection approaches are trained on adjusted datasets with controlled contamination ratios and tested on the original version.

The AUC-ROC performance is reported in Fig. \ref{fig:robustness}. Generally, the performance of all the anomaly detectors downgrades with the increasing contamination ratio.
Nevertheless, DIF has relatively clear superiority and stronger robustness in most of the datasets.
The success of deep ensemble on out-of-distribution robustness has been proved in many recent studies \cite{liu2022nooverhead,angelo2021repulsive,nam2021diversity}, which partially explains why deep ensemble-based methods show better robustness w.r.t. anomaly contamination than IF-based methods. 
However, these competing methods still fail to provide consistently good robustness (e.g., eRDP, eREPEN, and eDSVDD on \textit{R8} and eRECON on \textit{Backdoor}). It is mainly due to two reasons: (i) their ensemble processes may suffer from the diversity problem, and (ii) their scoring strategies and training objectives rely on strong assumptions such as the distance concept that might not hold in some datasets. 
It is interesting to note that eREPEN shows an uptrend in \textit{Shuttle}. REPEN uses LeSiNN to estimate initial anomaly scores when generating triplet mini-batches. 
As LeSiNN is with good robustness on \textit{Shuttle}, this initial estimation can reliably obtain more anomaly examples as positive data when the contamination ratio increases, and thus the triplet learning process might benefit from the augmentation of the positive class.

\subsection{Significance of the Synergy between Random Representations and Random Partition-based Isolation}\label{sec:significance}

We respectively replace random representations and random isolation-based anomaly scoring in DIF to investigate their synergy effect.

\subsubsection{Representation Scheme}\label{subsec:rep_scheme}
This experiment evaluates the effectiveness of our novel representation scheme by comparing it with representations produced by optimised neural networks, including RDP \cite{wang2021rdp}, REPEN \cite{pang2018Learning}, DSVDD \cite{ruff2018dsvdd} and Reconstruction-based Autoencoder \cite{aggarwal2017outlieranalysis}. 
That is, we replace random representations with representations learned by one of these methods, with all the other components of DIF fixed. These four variants are denoted as RDP-DIF, REPEN-DIF, DSVDD-DIF and RECON-DIF. 
RDP, REPEN, and DSVDD are originally designed for learning a good representation for anomaly detection, and a dense representation can be implicitly derived from Autoencoders.
The representations are evaluated by the AUC-ROC performance and the individual representation quality as follows.

\begin{figure}[t]
	\centering
	\includegraphics[width=0.47\textwidth]{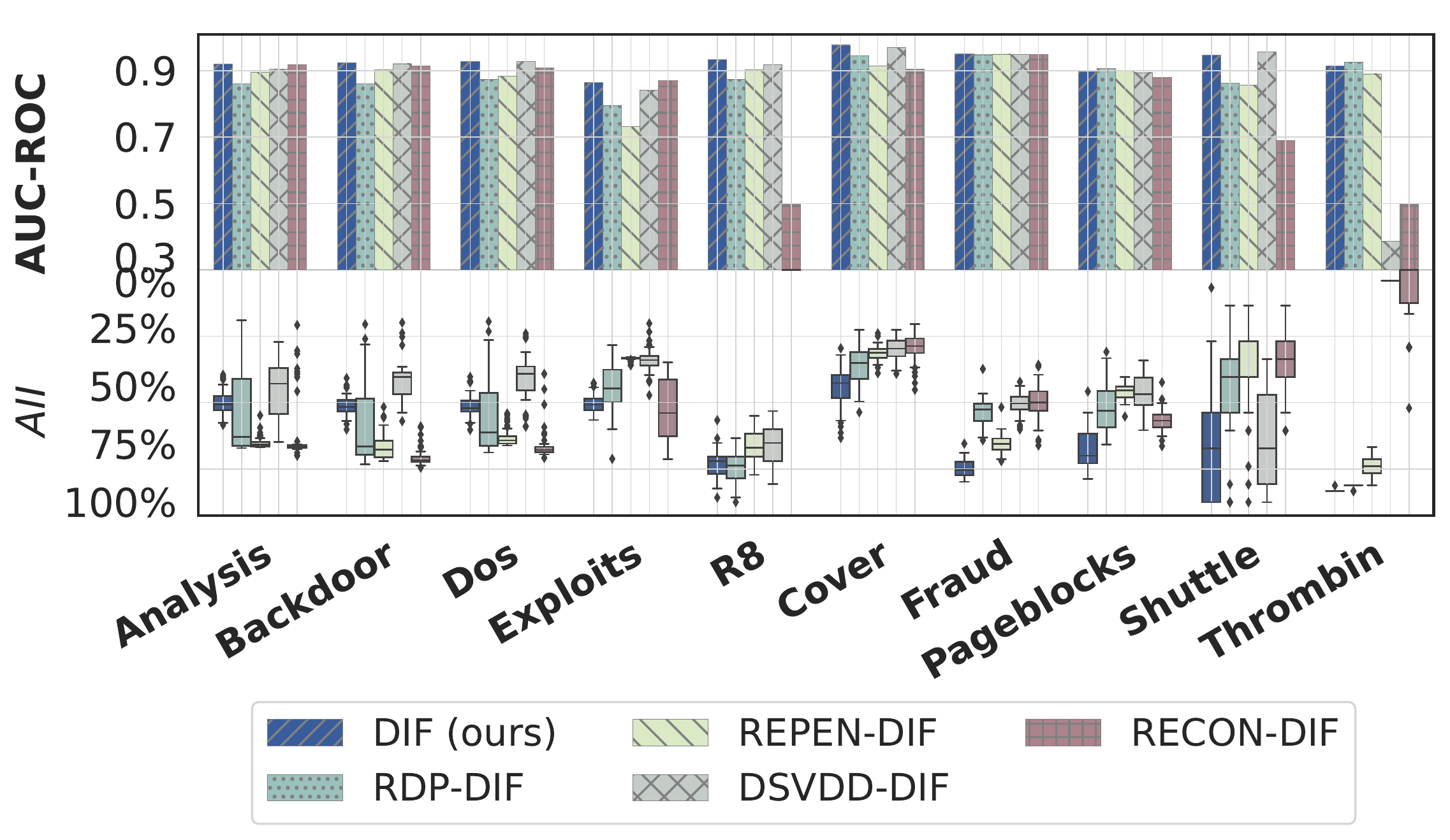}
	\caption{(\textbf{Top}) AUC-ROC of DIF and its variants that use optimised representations, and (\textbf{Bottom}) the quality (measured by $AII$) distribution of randomised representations used in DIF and representations optimised by RDP, REPEN, DSVDD, and RECON.}
	\label{fig:random}
\end{figure}

\textbf{AUC-ROC Results}. 
The AUC-ROC results are shown in the upper half of Fig. \ref{fig:random}. DIF outperforms four optimised representation ensemble-based methods on five datasets and shows very competitive performance to the best performer on the other five datasets. 
Averagely, the random representation ensemble contributes to 5\%, 5\%, 7\%, and 15\% AUC-ROC improvement than RDP-DIF, REPEN-DIF, DSVDD-DIF, and RECON-DIF, respectively. 
Although DIF may not be the best performer on all the datasets, it is very encouraging to see this performance achieved by the ensemble of random representations when compared to those optimised representations trained by state-of-the-art learning objectives.

\textbf{Quality per Representation}. To further analyse the mechanism behind the above results, we directly evaluate the quality of each representation produced by DIF and its variants. 
The representation quality is measured by Anomaly Isoability Index ($AII$), as introduced in Sec. \ref{sec:metric}.
The $AII$ results of each representation used in five anomaly detection methods are shown in Fig. \ref{fig:random} (bottom), in which we use box plots to present the quality distribution of representations produced by 50 representations in the ensemble framework.

Based on the above experiment results, the following three remarks can be made.

\begin{itemize}[leftmargin=*]
	\item Our representation scheme achieves desired diversity and randomness,
	while simultaneously maintaining 
	stable expressiveness in each representation, enabling excellent synergy with the downstream isolation-based anomaly scoring mechanism. This is the main driving force behind the superior performance of DIF. 
	
	\item Optimised representations can be with consistently good quality on some datasets (e.g., near 80\% true anomalies are well isolated by RECON-DIF on \textit{Analysis}, \textit{Backdoor}, and \textit{DoS}), whereas the lack of diversity in representations downgrades the efficacy of this ensemble framework.
	
	\item Optimisation may even lead to worse representations on some datasets compared to random representations (e.g., \textit{R8}, \textit{Cover}, \textit{Pageblocks} and \textit{Thrombin}). This may be due to the fact that the underlying assumption (e.g., one-class assumption) in their learning objectives may not hold in those datasets.

\end{itemize}

\subsubsection{Scoring Strategy}

As shown in the prior experiment, random representations are with good diversity and stable quality, fostering a unique excellent synergy effect in downstream ensemble-based anomaly scoring. This section further justifies this intuition by investigating the effectiveness of combining random representations with other anomaly scoring methods, including probability-based anomaly scoring method ECOD \cite{li2022ecod}, distance-based method KNN \cite{ramaswamy2000knn}, and density-based method LOF \cite{breunig2000lof}. Each of these methods is used to replace the isolation-based scoring process, with all the other modules fixed. 
These variants are denoted as DIF-ECOD, DIF-KNN, and DIF-LOF, respectively. Similarly, we evaluate their AUC-ROC performance and the individual scoring quality as follows.

\begin{figure}[t]
\centering
\includegraphics[width=0.46\textwidth]{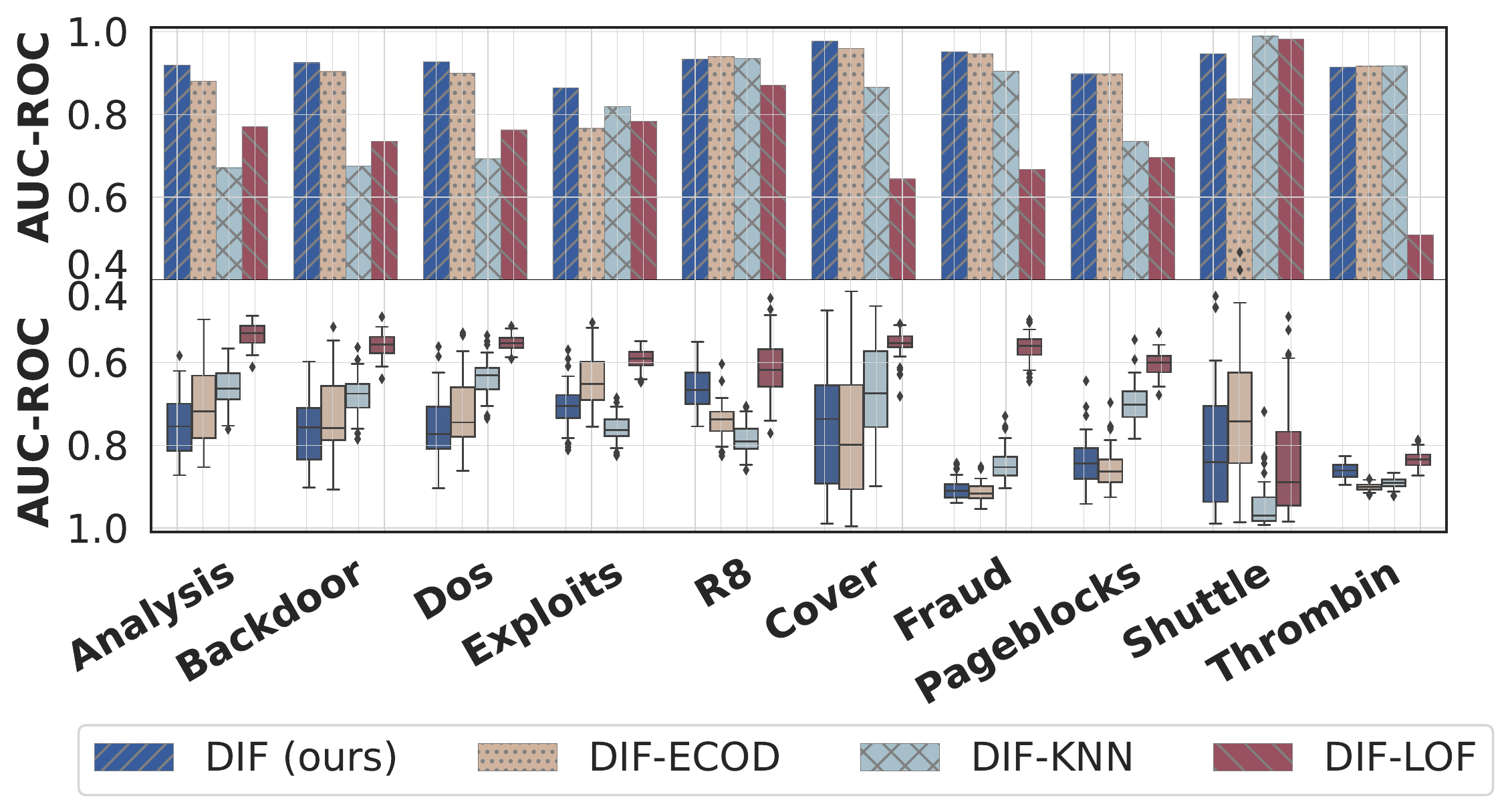}
\caption{(\textbf{Top}) AUC-ROC of DIF and its variants of using other anomaly scoring strategies upon random representations, and (\textbf{Bottom}) their effectiveness (measured by AUC-ROC) on each ensemble member within the randomised representation groups. 
}
\label{fig:whyif}
\end{figure}

\textbf{AUC-ROC Results}.
The AUC-ROC results are shown in the upper half of Fig. \ref{fig:whyif}. 
DIF outperforms these competing variants on seven out of ten datasets. 
Averagely, DIF outperforms DIF-ECOD, DIF-KNN, and DIF-LOF by 3\%, 13\%, and 25\%, respectively. 
The superiority further justifies the synergy effect in DIF. 
Note that KNN and LOF are with very heavy computational overhead. They take around 60 hours to handle large-scale datasets \textit{Cover} and \textit{Fraud}. 

\textbf{Quality per Anomaly Scoring Result.}
We also study the quality of individual anomaly scoring results produced by DIF and its variants on each random representation, as shown in Fig. \ref{fig:whyif} (bottom). The quality is also estimated by AUC-ROC here.

We make the following three remarks.

\begin{itemize}[leftmargin=*]
\item These competing scoring methods can produce markedly better individual scoring results than our isolation-based scoring mechanism on \textit{Exploits}, \textit{R8}, and \textit{Thrombin}. However, they only yield less effective or slightly better integrated results compared to DIF. 
These scoring methods are unable to leverage the diversity embedded in our representation scheme.

\item By contrast, DIF combines data representation and anomaly scoring in a successful unified ensemble learning framework. DIF achieves better integrated performance by fully leveraging the diversity and randomness of representations, e.g., above 0.9 AUC-ROC on \textit{R8} with the maximum individual value only achieving around 0.7. 

\item DIF is inferior to its variants on \textit{Shuttle}. It might be because anomalies in this dataset can be more easily identified by using the prior concepts used in these competing scoring methods (i.e., probability, distance, or density). However, these prior concepts may fail to work properly across all the datasets. 

\end{itemize}

\subsection{Ablation Study on CERE and DEAS}\label{sec:ablation}

Considering detection effectiveness and computational efficiency, we propose the Deviation-Enhanced Anomaly Scoring function (DEAS) and the Computation-Efficient deep Representation Ensemble method (CERE) in the specific implementation of DIF. This experiment is conducted to examine whether DIF can have better detection performance and use less training time with the help of DEAS and CERE. Two ablated variants are employed, i.e., \textbf{w/o} CERE replaces $\mathscr{G}_{\text{CERE}}$ with the conventional sequential ensemble process, and \textbf{w/o} DEAS uses the standard scoring function used in iForest to replace our scoring function $\mathscr{F}_\text{DEAS}$. 

The AUC-ROC and AUC-PR results of DIF and \textbf{w/o} DEAS are shown in Table \ref{tab:ablation}. 
DIF significantly outperforms \textbf{w/o} DEAS at the 90\% confidence interval and achieves approximate 11\% AUC-PR improvement. 
Besides, DIF costs considerably less training time with the help of CERE. The total training time across all the ten datasets is only approximate one-tenth of the variant \textbf{w/o} CERE. 
The AUC-ROC/AUC-PR results of \textbf{w/o} CERE are on par with DIF, which are omitted due to the space limitation. 
Based on the above comparison results, the contribution of the proposed DEAS and CERE is validated and quantitatively measured.

\begin{table}[t]
\centering
\caption{AUC-ROC and AUC-PR results of DIF and \textbf{w/o} DEAS, and training time (in seconds) of DIF and \textbf{w/o} CERE. \textbf{w/o} DEAS is an ablated version by replacing DEAS with the standard scoring function. \textbf{w/o} CERE only uses the typical method to produce the representation ensemble.  }
\scalebox{0.86}{
\begin{tabular}{p{1.3cm} |
		p{0.5cm}<{\centering}
		p{1.5cm}<{\centering} |
		p{0.5cm}<{\centering}
		p{1.5cm}<{\centering} |
		p{0.5cm}<{\centering}p{1.4cm}<{\centering} }
	\hline
	\multirow{2}[0]{*}{\textbf{Data}} & \multicolumn{2}{c|}{\textbf{AUC-ROC}} & \multicolumn{2}{c|}{\textbf{AUC-PR}} & \multicolumn{2}{c}{\textbf{Time (in seconds)}} \\
	\cline{2-7}
	& \textbf{DIF} & \textbf{w/o} DEAS & \textbf{DIF} & \textbf{w/o} DEAS & \textbf{DIF} & \textbf{w/o} CERE \\
	\hline
	
	Analysis & \textbf{0.931} & 0.922$_{\pm0.007}$ & \textbf{0.404} & 0.315$_{\pm0.058}$ & 10.5  & 69.2  \\
	Backdoor & \textbf{0.918} & 0.914$_{\pm0.006}$ & \textbf{0.453} & 0.381$_{\pm0.067}$ & 10.5  & 66.2  \\
	DoS   & \textbf{0.932} & 0.926$_{\pm0.007}$ & \textbf{0.440} & 0.388$_{\pm0.074}$ & 10.4  & 65.6  \\
	Exploits & \textbf{0.858} & 0.854$_{\pm0.008}$ & \textbf{0.273} & 0.248$_{\pm0.041}$ & 10.4  & 65.8  \\
	R8    & \textbf{0.930} & 0.915$_{\pm0.012}$ & \textbf{0.145} & 0.123$_{\pm0.034}$ & 5.2   & 19.2  \\
	Cover & \textbf{0.972} & 0.964$_{\pm0.014}$ & \textbf{0.246} & 0.210$_{\pm0.084}$ & 33.0  & 154.3  \\
	Fraud & \textbf{0.953} & 0.952$_{\pm0.002}$ & \textbf{0.387} & \textbf{0.387$_{\pm0.042}$} & 33.6  & 169.1  \\
	Pageblocks & 0.903 & \textbf{0.912$_{\pm0.007}$} & 0.547 & \textbf{0.576$_{\pm0.024}$} & 0.8   & 3.5  \\
	Shuttle & \textbf{0.941} & 0.923$_{\pm0.012}$ & \textbf{0.150} & 0.103$_{\pm0.019}$ & 0.5   & 1.0  \\
	Thrombin & 0.913 & \textbf{0.916$_{\pm0.002}$} & \textbf{0.468} & 0.448$_{\pm0.020}$ & 13.8  & 336.7  \\
	\hline
	\textit{Avg./Total} & \textbf{0.925} & 0.920$_{\pm0.008}$ & \textbf{0.351} & 0.318$_{\pm0.046}$ & 128.7  & 950.6  \\
	
	\textit{p-value} &   -    & 0.07  &   -    & 0.03  &    -   &  - \\
	
	\hline
\end{tabular}
}%
\label{tab:ablation}%
\end{table}%

\section{Conclusions}
This paper introduces DIF, a novel extension of iForest. DIF takes the deep neural network-based random representation ensemble as a new representation scheme, enabling significantly versatile data partition in diverse random directions on subspaces of different sizes.
The anomaly scoring can be facilitated by the synergy between random representations and random partition-based isolation. 
This enables DIF to fulfil (i) more effective isolation of anomalies, especially hard anomalies in data with intractable sparsity and non-linearity; 
(ii) liberation of isolation process from existing constraints to tackle the artefact problem;
and (iii) versatile ability to handle different data types.
Extensive experiments show that DIF significantly outperforms iForest and its existing extensions not only on tabular data but also on graph and time-series data. DIF also shows promising improvement compared to the ensemble of state-of-the-art deep anomaly detectors. 

In our future work, we plan to devise new scoring strategies, together with relevant neural network backbones, to address other challenging yet important anomaly detection tasks, e.g., identify possible abnormal subsets \cite{huang2021hybrid} and multi-view data \cite{ji2019multi}.

\ifCLASSOPTIONcompsoc
\section*{Acknowledgments}
\else
\section*{Acknowledgment}
\fi

Hongzuo Xu, Yijie Wang, and Yongjun Wang are supported by the National Key R\&D Program of China (No. 2022ZD0115302), the National Natural Science Foundation of China (No.61379052), the Science Foundation of Ministry of Education of China (No.2018A02002), the Postgraduate Scientific Research Innovation Project of Hunan Province (CX20210049), the Natural Science Foundation for Distinguished Young Scholars of Hunan Province (No.14JJ1026). Guansong Pang is supported in part by the Singapore Ministry of Education
(MOE) Academic Research Fund (AcRF) Tier 1 grant (21SISSMU031).

We thank Eamonn Keogh and his students for creating the UCR  time series anomaly archive that has been used in our experiment. We also thank the referees for their comments, which helped improve this paper considerably."

\ifCLASSOPTIONcaptionsoff
\newpage
\fi


\bibliographystyle{IEEEtran}
\bibliography{ref}

%

\begin{IEEEbiography}[{\includegraphics[width=1in,height=1.25in,clip,keepaspectratio]{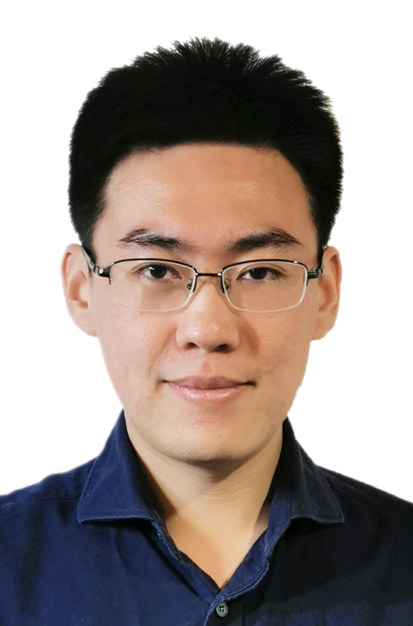}}]{Hongzuo Xu}
received the bachelor’s and master’s degree from the National University of Defense Technology, China, in 2017 and 2019, where he is currently pursuing the Ph.D. degree in computer science.
His research interests include anomaly detection, outlier interpretation, weakly-supervised learning
and data mining. 
\end{IEEEbiography}

\begin{IEEEbiography}[{\includegraphics[width=1in,height=1.25in,clip,keepaspectratio]{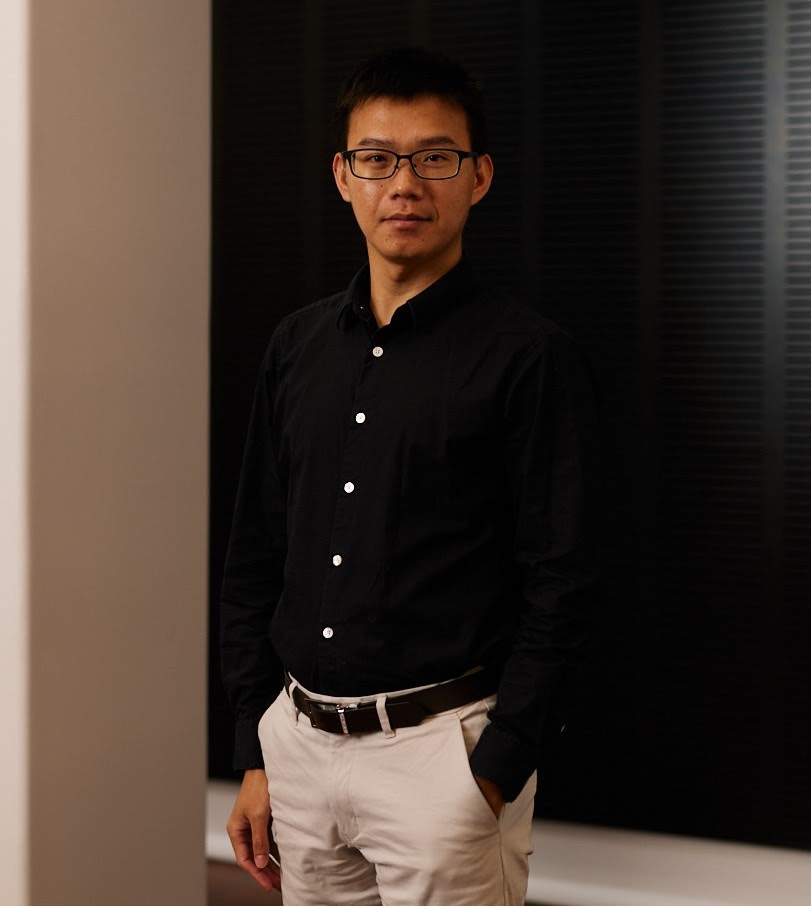}}]{Guansong Pang} is a tenure-track Assistant Professor of Computer Science in the School of Computing and Information Systems at Singapore Management University (SMU), Singapore. Before joining SMU, he was a Research Fellow with the Australian Institute for Machine Learning (AIML). He received a PhD degree from University of Technology Sydney in 2019. His research explores novel data mining and machine learning techniques and their applications, with a focus on detecting abnormal and unknown data.
\end{IEEEbiography}

\begin{IEEEbiography}[{\includegraphics[width=1in,height=1.25in,clip,keepaspectratio]{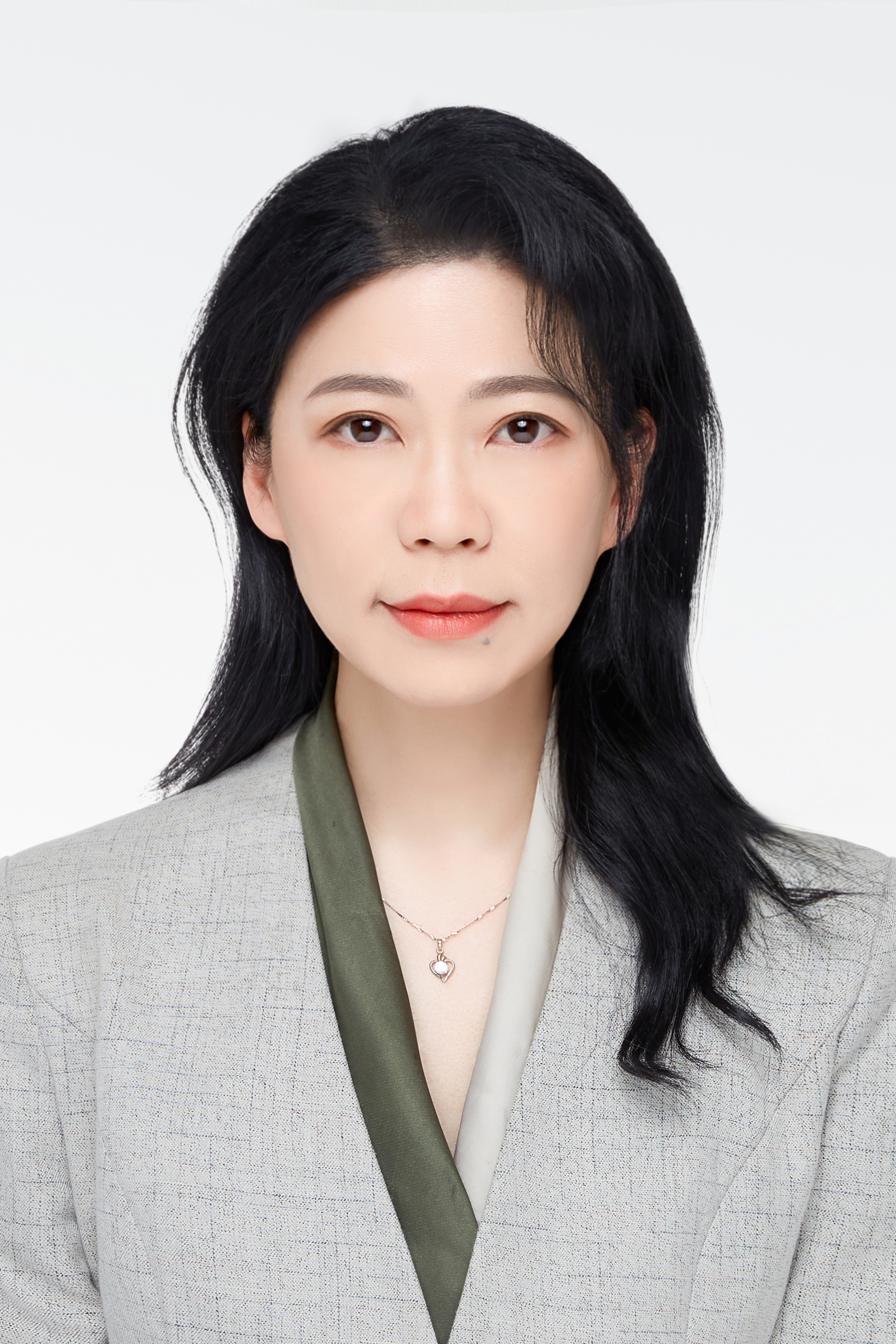}}]{Yijie Wang}
received the PhD degree in computer science and technology from the National University of Defense Technology in 1998. 
She was awarded the prize of National Excellent
Doctoral Dissertation by Ministry
of Education of PR China.
She is currently a Full Professor with the Science and Technology on Parallel and Distributed Processing Laboratory (PDL), National University of Defense Technology. Her research interests include
big data analysis, artificial intelligence and parallel and distributed processing.
\end{IEEEbiography}

\begin{IEEEbiography}[{\includegraphics[width=1in,height=1.25in,clip,keepaspectratio]{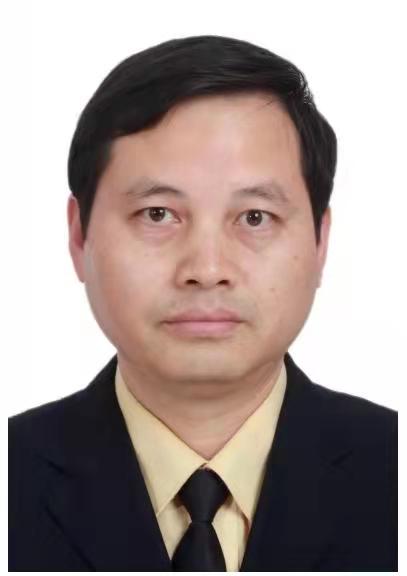}}]{Yongjun Wang}
received the Ph.D. degree in computer architecture from the National University of Defense Technology, China, in 1998. He is currently a Full Professor with the College of Computer, National University of Defense Technology, Changsha, China. His research interests include network security and system security.
\end{IEEEbiography}







\end{document}